\documentclass[11pt]{article}
\usepackage{tabularx}
\usepackage{array}
\usepackage[final]{acl} % camera-ready: final mode (no line numbers/anonymization)
\usepackage{times}
\usepackage{latexsym}
\usepackage{tcolorbox}
\usepackage[T1]{fontenc}
\usepackage[utf8]{inputenc}
\usepackage{microtype}
\usepackage{inconsolata}
\usepackage{amsmath}
\usepackage{booktabs}
\usepackage{multirow}
\usepackage{xcolor}
\usepackage{url}
\usepackage{graphicx}
\usepackage{float}
\usepackage{makecell}
\usepackage{listings}
\usepackage{placeins}
\usepackage{ragged2e}
\usepackage{fontawesome5}

\usepackage{xcolor}
\usepackage{listings}
\tcbuselibrary{listings,breakable,skins}

\lstdefinestyle{appendixjson}{
  basicstyle=\ttfamily\scriptsize,
  columns=fullflexible,
  breaklines=true,
  breakatwhitespace=false,
  keepspaces=true,
  showstringspaces=false,
  tabsize=2,
  upquote=true
}

\newtcblisting{jsonlistingbox}[1]{
  enhanced,
  breakable,
  title={#1},
  colback=black!2,
  colframe=black!55,
  colbacktitle=black!8,
  coltitle=black,
  fonttitle=\bfseries,
  boxrule=0.5pt,
  arc=0.8mm,
  left=1mm,
  right=1mm,
  top=1mm,
  bottom=1mm,
  listing only,
  listing engine=listings,
  listing options={
    style=appendixjson
  }
}

\newtcolorbox{promptbox}[1]{
  enhanced,
  breakable,
  title={#1},
  colback=black!2,
  colframe=black!55,
  colbacktitle=black!8,
  coltitle=black,
  fonttitle=\bfseries,
  boxrule=0.5pt,
  arc=0.8mm,
  left=1.2mm,
  right=1.2mm,
  top=1mm,
  bottom=1mm
}

\newcommand{\benchmark}{\textsc{EvoGenUI-Bench}}
\newcommand{\genui}{\textsc{EvoGenUI}}

\title{\benchmark: Evaluating LLMs as Multi-Turn Generative UI Assistants}

\author{
\textbf{Yue Peng},
\textbf{Lanke Xia}\textsuperscript{*},
\textbf{Zihan Wang}\textsuperscript{*},
\textbf{Jiahao Ye}\textsuperscript{*},
\textbf{Ke Ning},
\textbf{Hongyi Wen}\textsuperscript{\dag}
\\
New York University Shanghai
\\
\faGithub\ Code: \url{https://github.com/MAPS-research/EvoGenUI-Bench}
}

\begin{document}
\maketitle
\def\thefootnote{*}\footnotetext{Equal Contribution}
\def\thefootnote{\dag}\footnotetext{Correspondence to <hongyi.wen@nyu.edu>}
\renewcommand{\thefootnote}{\arabic{footnote}}\setcounter{footnote}{0}

\begin{abstract}

Large language models can generate interactive web interfaces, but reliable generative UI requires maintaining an executable artifact as user requests evolve. We introduce \benchmark{}, a benchmark for multi-turn interface maintenance comprising 150 five-turn tasks and 750 turns across three scenarios: information presentation, executable interaction, and tool-grounded external state. We execute generated artifacts in a browser and evaluate them using screenshots, source and DOM evidence, actor traces, and runtime logs. Beyond turn-level and episode-level success, we measure cross-turn retention with Adjacent Pass
Retention. Across eight models, even the strongest achieves 74.9\% Turn Pass while completing only 37.3\% of five-turn episodes; APR further falls to 52.4\% on tool-grounded tasks. Diagnostic analysis shows that presentation failures center on information architecture, interaction failures on derived-state propagation and affordance binding, and tool-grounded failures additionally involve external-state grounding and requirement decomposition. These results reframe generative UI evaluation from judging isolated outputs to testing whether interface behavior, derived state, external state, and assistant claims remain synchronized as the artifact evolves.

% Large language models can generate interactive interfaces, but it remains unclear whether they can reliably maintain these interfaces as user requirements evolve.
% We introduce \benchmark{}, a benchmark for multi-turn executable interface maintenance comprising 150 tasks and 750
% turns across three scenarios: information presentation, executable interaction, and tool-grounded external state. 
% Each task begins with an interface-generation request followed by four successive revisions. We execute generated artifacts in a browser and evaluate them using evidence-grounded measures of Presentation, Execution, and Alignment. In addition to turn-level and episode-level success, we measure cross-turn retention with Adjacent Pass Retention (APR), which captures how often a passing interface continues to pass after the next revision. Experiments reveal a persistent gap between turn-level correctness and cross-turn reliability: models can satisfy the current revision, yet reliability drops sharply when they must preserve executable behavior and text--artifact alignment across subsequent turns, revealing a substantial gap between producing locally successful updates and sustaining correct interface behavior over time, with distinct failure patterns across scenario types.
\end{abstract}

\section{Introduction}

\begin{figure}[t!]
    \centering
    \includegraphics[width=\columnwidth]{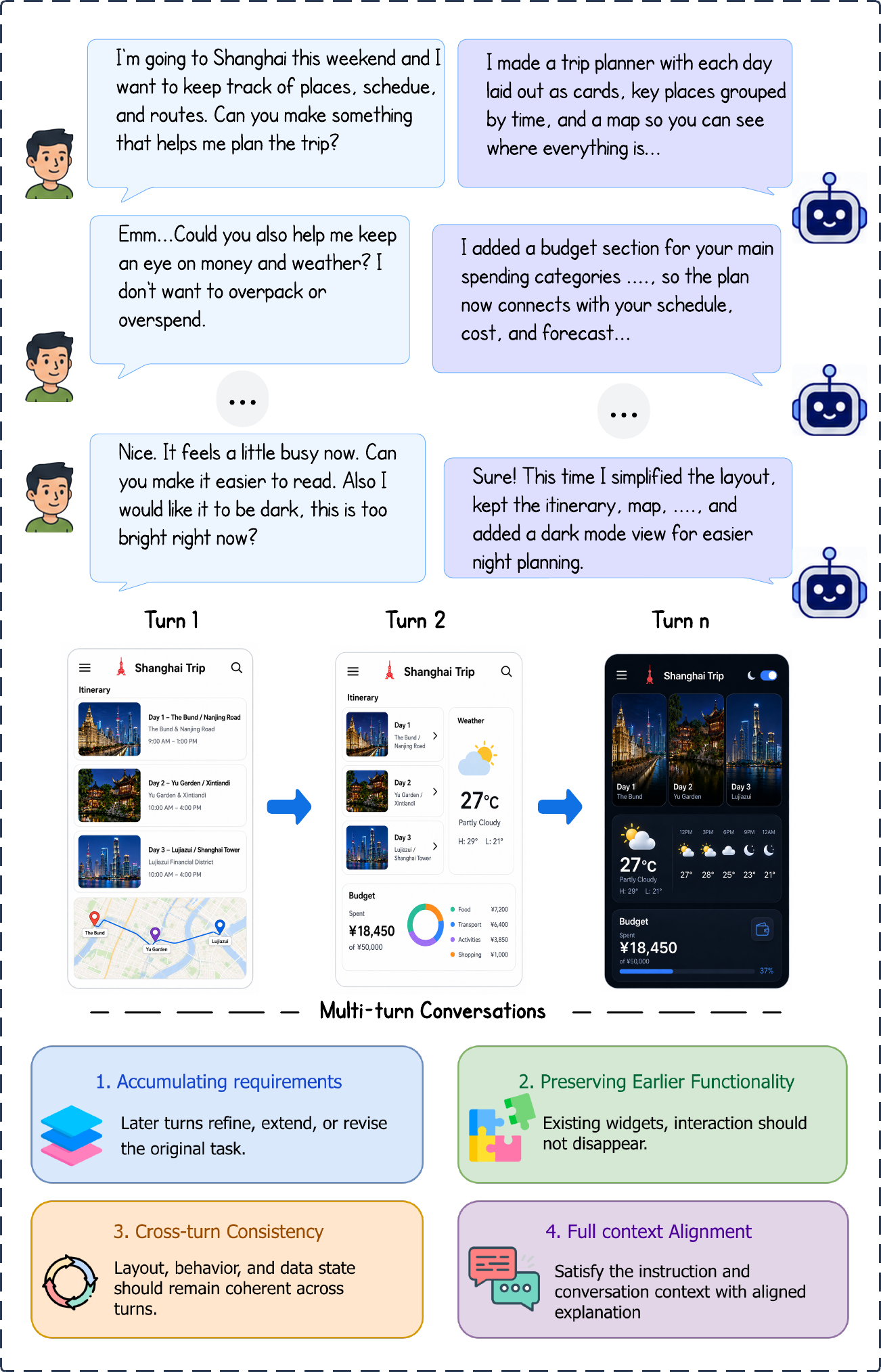}
    \caption{\textbf{Multi-turn artifact maintenance.} Rather than generating independent pages, the model repeatedly updates a single executable interface as user requirements evolve. Each update must implement the current request, preserve still-valid requirements, and keep interface behavior, visible state, and the accompanying response consistent with the conversation history.}
    \label{fig:multi-turn-conversation}
\end{figure}

Large language models (LLMs) can generate task-specific interactive interfaces, including dashboards, forms,
comparison views, and mini-applications. Recent work studies these interfaces as a direct response modality for
LLM-based assistants \citep{leviathan2026generativeui, chen2025generativeinterfaces, zhang2026miniappbench}. We
investigate whether LLMs can maintain an executable interface as user requirements evolve over multiple turns.
Figure~\ref{fig:multi-turn-conversation} illustrates a conversation in which the user adds filters and views, revises constraints, and extends a workflow while expecting earlier functionality to remain intact. We refer to this setting as \genui{}: across turns, the model updates a single executable interface to satisfy each new request without violating requirements that remain valid. Single-shot evaluation does not capture the nuances in this setting. 

Existing benchmarks evaluate frontend generation and revision, tool use, and agents operating within existing web environments \citep{zhu2025frontendbench,wu2026frontalk,yao2024taubench,zhou2024webarena}. 
These settings largely separate artifact construction from task execution: the model either develops an interface or operates through a pre-defined environment. Generative UI collapses this distinction: the model continually revises the interface through which the task itself is performed. Reliability must therefore be assessed over the full revision sequence, including whether behavior, state, and prior requirements remain consistent as the interface evolves.

We introduce \benchmark{}, a benchmark comprising 150 tasks and 750 turns across three scenario suites. Presentation-focused tasks evaluate the organization of dense information; interaction-focused tasks evaluate executable interfaces with local state; and tool-grounded tasks evaluate interfaces that mediate external runtime state. Each task consists of an initial interface-generation request followed by four successive revisions to the same interface.

For every requested artifact, we attempt to build and execute the generated interface in a browser. Buildable artifacts are scored for Presentation, Execution, and Alignment using the multi-surface evidence bundle defined in Section~\ref{sec:generation-execution}, while artifacts that cannot be built or executed are counted as failures. We measure turn-level success with Turn Pass, complete-episode success with TP@5, and cross-turn retention with Adjacent Pass Retention (APR), the share of qualifying next turns that pass given a passing previous turn. The best-performing model reaches 74.9\% overall Turn Pass but completes only 37.3\% of five-turn episodes. After pooling all such adjacent pairs across the eight models, tool-grounded APR is 52.4\%. Failure mechanisms also differ substantially across suites.

% For every requested artifact, we attempt to build and execute the generated interface in a browser. Buildable artifacts are scored for Presentation, Execution, and Alignment using the multi-surface evidence bundle defined in Section~\ref{sec:generation-execution}, while artifacts that cannot be built or executed are counted as failures. We measure turn-level success with Turn Pass, complete-episode success with TP@5, and cross-turn retention with Adjacent Pass Retention (APR), the share of qualifying next turns that pass given a passing previous turn. TP@5 is the percentage of episodes in which all five turns pass. The best-performing model reaches 74.9\% overall Turn Pass but completes only 37.3\% of five-turn episodes. After pooling all such adjacent pairs across the eight models, tool-grounded APR is 52.4\%. Failure mechanisms also differ substantially across suites.

Our contributions are: (1) \benchmark{}, a multi-turn benchmark for evaluating the maintenance of executable interfaces across cumulative revisions; (2) a browser-based, evidence-grounded evaluation protocol that combines visual, behavioral, source-level, and runtime evidence with turn-, episode-, and transition-level reliability metrics; (3) a systematic evaluation of eight recent models spanning multiple families, revealing a substantial gap between turn-level success and sustained reliability; and (4) a human-validated diagnostic taxonomy that identifies distinct mechanisms of interface-maintenance failure across presentation, interaction, and tool-grounded scenarios.

\section{Related Work}
\subsection{Generative Interfaces and Iterative Development}

Existing Generative UI benchmarks assess visual fidelity \citep{si-etal-2025-design2code}, interactive behavior and executable frontend generation \citep{xiao2025interaction2code,zhu2025frontendbench,sun2025fullfront}, and multi-file website generation \citep{lu2025webgenbenchevaluatingllmsgenerating}. MiniAppBench extends this line of work to principle-driven, interaction-intensive mini-applications evaluated through static inspection and agent-based exploration \citep{zhang2026miniappbench}. These benchmarks primarily focus on self-contained generation episodes rather than the evolution of a persistent interface under successive user requests.

Recent work has begun to examine generation under evolving requirements. FronTalk studies multi-turn frontend development with textual and visual feedback, identifying the forgetting or overwriting of prior features as a central challenge \citep{wu2026frontalk}. SlopCodeBench shows that iterative code extension can satisfy intermediate checkpoints while accumulating structural degradation \citep{slopcodebench2026}, and MultiChallenge documents broader failures in realistic multi-turn instruction following \citep{sirdeshmukh2025multichallengerealisticmultiturnconversation}. \benchmark{} extends this direction by treating the generated interface as both a persistent assistant output and the interaction layer. Each revision must incorporate the latest request while preserving prior requirements, executable behavior, and consistency with the assistant response; tool-grounded tasks additionally require synchronization with external runtime state.

\subsection{Interactive Agents and Stateful Tool Use}

WebArena and VisualWebArena evaluate agents operating websites, while OSWorld and AgentBench extend this paradigm to open-ended computer environments and broader interactive tasks \citep{zhou2024webarena,koh-etal-2024-visualwebarena,xie2024osworld,liu2024agentbench}. In these benchmarks, the environment supplies the interface through which the agent acts. In \benchmark{}, the model instead generates and continually revises that interface.

ToolLLM and the Berkeley Function Calling Leaderboard evaluate tool selection and invocation \citep{qin2024toolllm,patil2025bfcl}, whereas $\tau$-bench, ToolSandbox, and AppWorld introduce environments in which tool calls modify persistent state \citep{yao2024taubench,lu2025toolsandbox,trivedi-etal-2024-appworld}. These benchmarks primarily assess whether an agent selects appropriate tools and produces the intended state changes. The tool-grounded suite of \benchmark{} asks whether those operations are also exposed correctly through an evolving, user-operable interface, maintaining consistency among visible controls, browser interactions, local UI state, and authoritative external state.

\subsection{Evaluation of Executable Interactive Artifacts}

Open-ended interfaces cannot be evaluated reliably through reference renderings, static code analysis, or text-only judgments alone. SWE-bench evaluates generated code through execution in a software environment, and frontend benchmarks similarly use sandboxed builds, browser execution, and functional tests \citep{jimenez2024swebench,zhu2025frontendbench}. ArtifactsBench complements execution with temporal screenshots and multimodal judging to capture visual and interactive behavior \citep{zhang2025artifactsbenchbridgingvisualinteractivegap}.

LLM-as-a-judge methods such as MT-Bench, AlpacaEval, and Prometheus provide scalable, rubric-guided evaluation of open-ended outputs while motivating careful control of evaluator evidence and bias \citep{zheng2023mtbench,dubois2024alpacaeval,kim2024prometheus}. Executable interface maintenance requires these judgment-based methods to be grounded in execution evidence: an artifact may build successfully while exposing broken controls, retaining stale state, losing prior functionality, or contradicting the assistant response. \benchmark{} therefore combines rendered UI observations, source and DOM evidence, browser interactions, runtime traces, textual responses, and, when applicable, authoritative external-state readback within the shared evidence bundle.

\section{\benchmark}
\begin{figure*}[t!]
    \centering
    \includegraphics[width=0.86\textwidth]{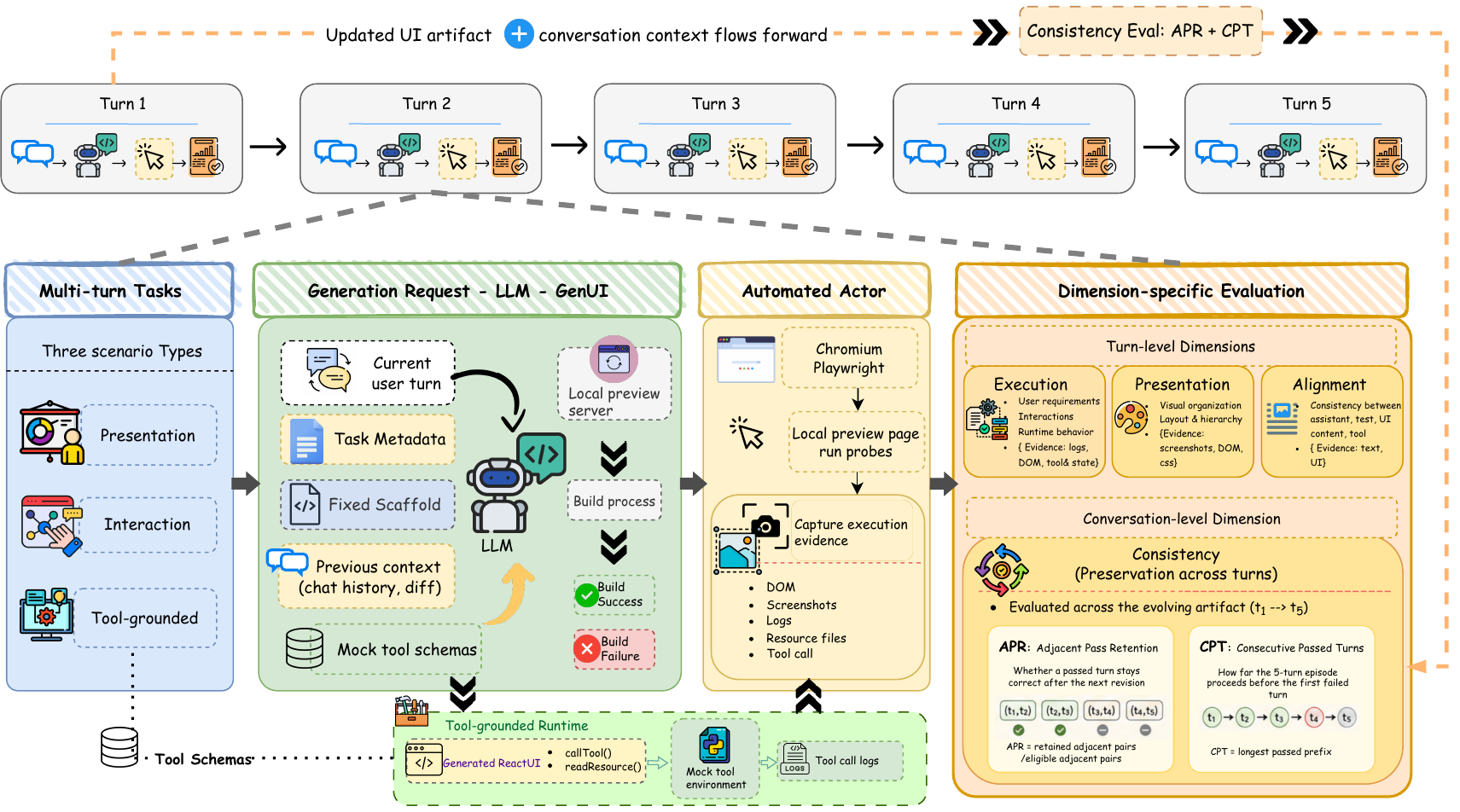}
    \caption{\textbf{Evaluation pipeline.} \benchmark{} generates an updated interface, executes it in a browser,
    collects multi-surface evidence, and computes turn-level scores and derived statistical reliability metrics.}
    \label{fig:evaluation-pipeline}
\end{figure*}

\subsection{Task Formulation}
\label{sec:task-formulation}

We formulate the evaluation of generative UIs as multi-turn artifact maintenance. Each task is a five-turn episode in which the user successively revises the same web interface rather than requesting independent pages. At each turn, the model receives the current request, public task context, optional public tool contracts, compact evidence from earlier turns, and the most recently generated source code. It must return both a user-facing response and the complete source code for the updated interface.

The task is cumulative: a valid update must implement the new request while preserving all still-valid prior requirements. The evaluation unit is therefore an evolving executable artifact rather than an isolated page. Its behavior, visible state, and accompanying assistant response must remain mutually consistent across revisions. Section~\ref{sec:generation-execution} describes how the benchmark harness executes each update and collects evaluation evidence.

\subsection{Design and Construction}
\label{sec:task-construction}
To prevent evaluation leakage, the benchmark separates generator-visible requests, context, and tool contracts from private validation requirements, backend state, actor-only information, and scoring criteria. Each private requirement identifies at least one observable evidence surface: the rendered UI, DOM, interaction trace, source code, assistant response, tool log, or runtime state. Validation targets the requested semantics rather than a reference implementation, allowing functionally equivalent solutions to pass while rejecting static shells and unsupported claims about interface or external state.

All tasks are human-authored to maintain a fixed five-turn revision structure, balanced coverage across scenario types, and, where applicable, deterministic external state. Machine-readable domain metadata guides task sampling and diversity auditing. The benchmark contains 150 distinct domain labels across 150 tasks, with each task pairing its domain with an interface type and a targeted challenge, such as evidence-board synchronization, state-machine simulation, linked-view propagation, or tool-backed status readback. This design reduces the risk of near-duplicate prompt templates while preserving a common evaluation structure.

A second reviewer checked each task for public-turn clarity, leakage of private validation criteria, meaningful cross-turn revision pressure, domain duplication, and requirement observability. Tasks that failed any check were revised. Appendix~\ref{app:task-construction} describes the complete construction and review workflow, and Appendix~\ref{app:example-task} presents a tool-grounded example.

\subsection{Benchmark Composition}
\label{sec:dataset-statistics}

The \benchmark{} suite contains 150 tasks and 750 requested turns, divided equally among three 50-task scenario suites: Presentation, Interaction, and Tool-grounded. Each task comprises five user turns and a private validation contract for every turn. Presentation tasks emphasize explanatory and decision-support artifacts; Interaction tasks cover stateful mini-applications and reasoning workbenches; and Tool-grounded tasks require interfaces that read, write, reconcile, or refresh external state. The suites contain equal numbers of tasks and turns but intentionally differ in validation-contract density, as shown in Table~\ref{tab:benchmark-overview}.

% The \benchmark{} suite contains 150 tasks and 750 requested turns, organized as three 50-task subsets:
% Presentation, Interaction, and Tool-grounded. Each task consists of five user turns and a private validation
% contract for every turn. Presentation emphasizes explanatory and decision-support artifacts, Interaction covers
% stateful mini-apps and reasoning workbenches, and Tool-grounded tasks require UIs that read, write, reconcile, or
% refresh external state. The subsets are balanced by task and turn count, not by validation-contract density
% (Table~\ref{tab:benchmark-overview}).

\begin{table}[ht]
\centering
\scriptsize
\setlength{\tabcolsep}{3.0pt}
\renewcommand{\arraystretch}{1.06}
\begin{tabular}{@{}lrrrr@{}}
\toprule
\textbf{Subset} & \textbf{Tasks} & \textbf{Domains} & \textbf{Tools/task} & \textbf{Req./turn} \\
\midrule
Presentation & 50 & 50 & 0 & 3.0 \\
Interaction & 50 & 50 & 0 & 5.9 \\
Tool-grounded & 50 & 50 & 6--22 & 11.9 \\
\midrule
\textbf{Total} & \textbf{150} & \textbf{150} & -- & \textbf{6.9} \\
\bottomrule
\end{tabular}
\caption{\textbf{Benchmark composition.} Every task contains five turns, yielding 750 requested turns in total. Req./turn denotes the mean number of private validation requirements per turn; these requirements are hidden from the generator.}
\label{tab:benchmark-overview}
\end{table}

\section{Evidence-Grounded Evaluation}
We evaluate each requested turn through interface generation, browser execution, evidence collection, and model-based scoring. Figure~\ref{fig:evaluation-pipeline} summarizes this workflow and the evidence passed between stages.

\subsection{Generation and Evidence Collection}
\label{sec:generation-execution}

At each turn, the benchmark harness builds the returned source code in a fixed web environment and serves the resulting interface in a browser. It records build and browser evidence, while an interaction actor operates the running interface to collect behavioral traces and observations. Execution proceeds sequentially: later turns depend on the previously generated source, collected snapshots, and, for tool-grounded tasks, restored runtime state.

If a required artifact is missing, downstream turns that depend on it are blocked rather than restarted from a clean state; the corresponding requested slots are not silently omitted from evaluation. Actor-reported interaction status is advisory rather than an official label. Instead, the recorded interaction evidence is passed to the evaluator and incorporated into the dimension-level scores described below.

\subsection{Evaluation Dimensions}
\label{sec:evaluation-dimensions}

% The model-based evaluator assigns three turn-level scores. \textbf{Presentation} covers the coherence, readability, and domain fit of the rendered UI. \textbf{Execution} covers implementation of the current request and relevant still-valid prior requirements as working interface behavior or visible, evidence-backed state. \textbf{Alignment} checks whether the assistant text, source, visible UI, interaction observations, and runtime logs describe the same behavior. Appendix~\ref{app:evaluation-rubrics} gives the full rubrics.

% The evaluator combines three evidence types: rendered state (the final DOM and screenshot), execution evidence
% (actor observations, tool and resource logs, and runtime snapshots), and artifact context (the source summary,
% build status, and compact previous-turn context). These inputs distinguish visual organization failures from broken
% controls, stale derived state, unsupported assistant claims, and tool-state contradictions.

The model-based evaluator assigns three turn-level scores. \textbf{Presentation} evaluates the coherence, readability, and domain appropriateness of the rendered interface. \textbf{Execution} evaluates whether the current request and all relevant, still-valid prior requirements are implemented as working interface behavior or visible, evidence-supported state. \textbf{Alignment} evaluates whether the assistant response, source code, rendered UI, interaction observations, and runtime logs are mutually consistent. Appendix~\ref{app:evaluation-rubrics} provides the complete scoring rubrics.

The evaluator combines three types of evidence: rendered state, comprising the final DOM and screenshot; execution evidence, comprising actor observations, tool and resource logs, and runtime snapshots; and artifact context, comprising the source summary, build status, and compact context from previous turns. Together, these inputs allow the evaluator to distinguish failures in visual organization from broken controls, stale derived state, unsupported assistant claims, and contradictions between the interface and tool-grounded state.

\subsection{Reliability Metrics}
\label{sec:metrics}

We report reliability at three levels: Turn Pass (\textbf{TP}) measures correctness at an individual turn; \textbf{TP@5}
and Consecutive Passed Turns (\textbf{CPT}) measure sustained success over an episode; and Adjacent Pass Retention
(\textbf{APR}) measures whether success persists across adjacent revisions. All four metrics derive from the same official turn-level decision. 

Formally, for turn $t$ of episode $\mathcal{T}$, let $p_{\mathcal{T},t}\in\{0,1\}$ denote the official pass indicator. It equals 1 only if generation succeeds, the returned source yields an evaluable artifact, and the Presentation, Execution, and Alignment scores are each at least 4. Turn Pass is the mean of these indicators over all requested model--turn slots; generation, build, and execution failures therefore remain failures under requested-slot accounting. 

For an $N$-turn episode $\mathcal{T}$,
\begin{equation}
P_{\mathrm{TP@5}}(\mathcal{T})=\prod_{t=1}^{N}p_{\mathcal{T},t},
\quad
\mathrm{CPT}(\mathcal{T})=\sum_{t=1}^{N}\prod_{j=1}^{t}p_{\mathcal{T},j}.
\label{eq:episode-metrics}
\end{equation}
%TP@5 is the first expression with $N=5$; CPT counts the initial run of passes and is reported in turns out of five.
Because every episode contains five turns, $P_{\mathrm{TP@5}}(\mathcal{T})$ indicates whether all five turns pass. CPT measures the length of the initial uninterrupted run of passing turns and ranges from zero to five. Suite-level TP@5 and CPT are obtained by averaging these episode-level quantities across tasks.

For comparison, we also report a diagnostic independence baseline for TP@5.
For each model, let $\hat{p}_t$ denote its empirical pass rate over all
requested slots at turn position $t$. If the five turn outcomes within an
episode were independent with these position-specific pass probabilities,
the expected TP@5 would be
\begin{equation}
P_{\mathrm{indep}}^{\mathrm{TP@5}}
=
\prod_{t=1}^{5}\hat{p}_t.
\label{eq:indep-5t}
\end{equation}

% TP, TP@5, and CPT use requested-slot accounting. APR instead conditions on an already passing artifact and
% therefore uses an eligible-transition denominator. 

Let $e_{\mathcal{T},t}=1$ if the generation call for turn $t$ returns a model response, and let $e_{\mathcal{T},t}=0$ if no response is obtained because of a provider failure. Invalidly formatted responses and build failures remain countable and are assigned $p_{\mathcal{T},t}=0$. For a set of episodes $\mathcal{S}$,
\begin{equation}
\mathrm{APR}(\mathcal{S})=
\frac{\sum_{\mathcal{T}\in\mathcal{S}}\sum_{t=2}^{N_{\mathcal{T}}}
e_{\mathcal{T},t}p_{\mathcal{T},t-1}p_{\mathcal{T},t}}
{\sum_{\mathcal{T}\in\mathcal{S}}\sum_{t=2}^{N_{\mathcal{T}}}
e_{\mathcal{T},t}p_{\mathcal{T},t-1}}.
\label{eq:apr}
\end{equation}
% APR therefore asks whether the next countable turn passes after a passing previous turn. TP, TP@5, and CPT
% are macro-averaged over tasks at suite level, whereas APR pools eligible adjacent transitions.

APR therefore measures the probability that turn $t$ passes given that turn $t-1$ passed and the generation call for turn $t$ returned a model response. It is an outcome-level retention measure: a failed transition may result from regression on a prior requirement, failure to satisfy the new request, or both. Appendix~\ref{app:apr-attribution-audit} separates attributable APR failures into cases involving regression on prior requirements and cases involving only the new requirement.

TP, TP@5, and CPT are macro-averaged across tasks at the suite level, whereas APR pools all eligible adjacent transitions. Appendix~\ref{app:evaluation-metrics} describes the countable-only diagnostics and bootstrap procedure. Table~\ref{tab:metric-uncertainty} reports task-level confidence intervals and independence-baseline values.

\subsection{Human Validation}
\label{sec:human-evaluation}

We validate automatic pass/fail decisions on a blinded sample of 240 turn-level cases covering all eight generator
models and all three suites, with 10 cases per model--suite pair and stratification by automatic pass/fail outcome.
Three non-author annotators independently label every case while blinded to model identity and the automatic
decision; majority vote defines the human reference. The automatic evaluator reaches 86.7\% accuracy and Cohen's
$\kappa=0.73$ against these labels, while inter-annotator agreement is Fleiss' $\kappa=0.78$.

% Avoid stretching run-in heading gaps on float-constrained experiment pages.
\raggedbottom
\section{Experiments}
\begin{table*}[t]
\centering
\scriptsize
\setlength{\tabcolsep}{1.9pt}
\renewcommand{\arraystretch}{1.08}
\begin{tabular}{@{}>{\raggedright\arraybackslash}p{0.125\textwidth}*{16}{>{\centering\arraybackslash}p{0.046\textwidth}}@{}}
\toprule
\multirow{2}{*}{\textbf{Model}}
& \multicolumn{4}{c}{\textbf{Presentation}}
& \multicolumn{4}{c}{\textbf{Interaction}}
& \multicolumn{4}{c}{\textbf{Tool-grounded}}
& \multicolumn{4}{c}{\textbf{Overall}} \\
\cmidrule(lr){2-5} \cmidrule(lr){6-9} \cmidrule(lr){10-13} \cmidrule(lr){14-17}
& \textbf{TP} & \textbf{TP@5} & \textbf{CPT} & \textbf{APR}
& \textbf{TP} & \textbf{TP@5} & \textbf{CPT} & \textbf{APR}
& \textbf{TP} & \textbf{TP@5} & \textbf{CPT} & \textbf{APR}
& \textbf{TP} & \textbf{TP@5} & \textbf{CPT} & \textbf{APR} \\
\midrule
Claude-Opus-4.7 & \textbf{82.4} & \textbf{52.0} & \textbf{3.28} & \textbf{91.3} & \textbf{81.2} & \textbf{40.0} & 3.02 & \textbf{86.8} & 61.2 & \textbf{20.0} & 2.14 & 68.6 & \textbf{74.9} & \textbf{37.3} & \textbf{2.81} & \textbf{83.6} \\
GPT-5.5 & 44.8 & 4.0 & 0.94 & 50.6 & 77.6 & \textbf{40.0} & \textbf{3.24} & 80.6 & \textbf{62.8} & \textbf{20.0} & \textbf{2.44} & \textbf{72.7} & 61.7 & 21.3 & 2.21 & 71.0 \\
Claude-4.5-Haiku & 76.0 & 32.0 & 3.08 & 82.8 & 56.0 & 8.0 & 1.36 & 60.9 & 21.2 & 0.0 & 0.40 & 20.0 & 51.1 & 13.3 & 1.61 & 65.7 \\
Qwen3.6-Plus & 60.0 & 12.0 & 1.76 & 72.3 & 46.8 & 10.0 & 1.58 & 64.1 & 25.2 & 0.0 & 0.62 & 34.6 & 44.0 & 7.3 & 1.32 & 61.7 \\
Gemini-3-Flash & 62.8 & 8.0 & 1.56 & 68.5 & 47.6 & 6.0 & 1.20 & 57.0 & 14.8 & 0.0 & 0.30 & 18.8 & 41.7 & 4.7 & 1.02 & 57.8 \\
Gemini-3.1-Pro & 38.4 & 8.0 & 1.14 & 70.3 & 25.2 & 12.0 & 0.78 & 79.6 & 7.2 & 0.0 & 0.16 & 29.4 & 23.6 & 6.7 & 0.69 & 68.6 \\
Qwen3-Coder & 44.0 & 6.0 & 1.44 & 53.5 & 21.2 & 0.0 & 0.72 & 32.0 & 4.4 & 0.0 & 0.14 & 0.0 & 23.2 & 2.0 & 0.77 & 42.2 \\
GLM-4.5-Air & 35.2 & 2.0 & 1.10 & 50.0 & 24.8 & 2.0 & 0.54 & 47.2 & 3.2 & 0.0 & 0.16 & 42.9 & 21.1 & 1.3 & 0.60 & 48.5 \\
\midrule
\textbf{Aggregate} & \textbf{55.5} & \textbf{15.5} & \textbf{1.79} & \textbf{71.1} & \textbf{47.6} & \textbf{14.8} & \textbf{1.56} & \textbf{68.7} & \textbf{25.0} & \textbf{5.0} & \textbf{0.80} & \textbf{52.4} & \textbf{42.7} & \textbf{11.8} & \textbf{1.38} & \textbf{66.5} \\
\bottomrule
\end{tabular}
\caption{\textbf{Main results.} Models are ordered by overall TP. TP is turn pass rate (\%); TP@5 is the percentage of episodes in which all five turns pass; CPT is the average initial run of passed turns; and APR is adjacent pass retention (\%). Overall averages the three suites for TP, TP@5, and CPT and pools eligible transitions for APR. The final row macro-averages model rows for TP, TP@5, and CPT and pools eligible transitions for APR. Bold marks column best and aggregate values.}
\label{tab:main-results}
\end{table*}

\subsection{Models}
\label{sec:models}

We evaluate eight models spanning several recent model families:
GPT-5.5 \citep{openai2026gpt55systemcard},
Qwen3.6-Plus \citep{alibabacloud2026qwen36plus},
Qwen3-Coder \citep{qwen2025qwen3coder},
GLM-4.5-Air \citep{zeng2025glm45},
Gemini-3-Flash \citep{googledeepmind2025gemini3flash},
Gemini-3.1-Pro \citep{googledeepmind2026gemini31pro},
Claude-4.5-Haiku \citep{anthropic2025claudehaiku45systemcard},
and Claude-Opus-4.7 \citep{anthropic2026claudeopus47systemcard}.
Appendix~\ref{app:evaluated-models} provides the exact provider identifiers and inference settings used for each model.

Where required, we use provider-specific transport or output modes while preserving a common normalized output contract: non-empty assistant text accompanied by the generated source files.

\subsection{Experimental Setup}
\label{sec:experimental-setup}

Each model is evaluated on the same 150 tasks, with five requested turns per task, using a common execution harness, evidence-collection protocol, and evaluator. Execution is sequential: each turn reuses the previously generated source and collected snapshots and, for tool-grounded tasks, the restored runtime state. For the main results, we use MiMo-V2.5 \citep{mimov25} as both the interaction actor and the evaluator. We use refreshed full-page screenshots and fixed evaluation rubrics across all evaluated models.

\subsection{Main Results}
\label{sec:main-results}

\begin{figure*}[t]
\centering
\begin{minipage}[t]{0.47\textwidth}
\centering
\includegraphics[width=\linewidth,height=0.18\textheight,keepaspectratio]{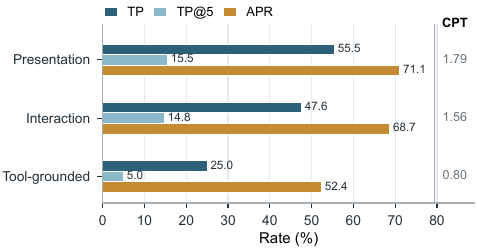}\par\smallskip
\small\textbf{(a) Suite difficulty}
\end{minipage}\hfill
\begin{minipage}[t]{0.47\textwidth}
\centering
\includegraphics[width=\linewidth,height=0.18\textheight,keepaspectratio]{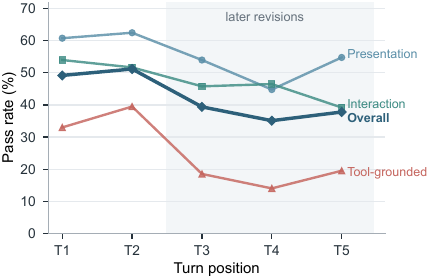}\par\smallskip
\small\textbf{(b) Turn position}
\end{minipage}
\caption{\textbf{Main empirical views.} (a) Turn Pass, TP@5, APR, and CPT by suite. (b) Pass rate over requested model--turn slots, with later revisions shaded.}
\label{fig:main-result-views}
\end{figure*}

\paragraph{Finding 1: Turn-level success overstates episode-level reliability.}
Claude-Opus-4.7 achieves the highest overall TP (74.9\%) and APR (83.6\%), yet only 37.3\% of its episodes pass all five turns. Averaged across models, TP is 42.7\%, whereas TP@5 is only 11.8\%. This gap shows that success on individual turns does not translate directly into reliable completion of an entire revision sequence.

% \paragraph{Finding 1: turn-level success overstates conversation reliability.}
% Claude-Opus-4.7 leads overall Turn Pass (74.9\%) and APR (83.6\%), but only 37.3\% of its episodes pass all five
% turns. Across model rows, mean Turn Pass is 42.7\% and mean TP@5 is 11.8\%. 

Observed TP@5 exceeds the independence estimate in
Equation~\ref{eq:indep-5t} for all eight models. Representative comparisons are 21.3\% versus 8.7\% for GPT-5.5, 7.3\% versus 1.3\% for Qwen3.6-Plus, 4.7\% versus 1.2\% for Gemini-3-Flash, and 37.3\% versus 23.4\% for Claude-Opus-4.7 (Table~\ref{tab:metric-uncertainty}). Passing turns therefore cluster within episodes, potentially because episodes differ in difficulty. Accordingly, TP@5 measures end-to-end workflow success rate.

Models also exhibit distinct suite-level profiles. GPT-5.5 and Claude-Opus-4.7 both achieve 40.0\% TP@5 on Interaction tasks. Qwen3.6-Plus performs substantially better on Presentation than on Tool-grounded tasks, with TP decreasing from 60.0\% to 25.2\%. Gemini-3.1-Pro has a low overall TP of 23.6\% but relatively high APR after reaching a passing state on Presentation (70.3\%) and Interaction (79.6\%). This contrast illustrates the conditional nature of APR: a model may preserve a correct artifact when one is reached even if it rarely reaches such a state.

% Across all eight models, observed TP@5 exceeds this baseline. Representative comparisons are 21.3\%
% versus 8.7\% for GPT-5.5, 7.3\% versus 1.3\% for Qwen3.6-Plus, 4.7\% versus 1.2\% for Gemini-3-Flash, and 37.3\%
% versus 23.4\% for Claude-Opus-4.7 (Table~\ref{tab:metric-uncertainty}). Passing turns therefore cluster within
% episodes. The comparison calibrates TP@5 as workflow survival, not as a causal measure of forgetting.

% Model profiles also differ across suites. GPT-5.5 and Claude-Opus-4.7 both reach 40.0\% TP@5 on
% interaction tasks. Qwen3.6-Plus is substantially stronger on presentation than tool-grounded tasks (60.0\% versus
% 25.2\% Turn Pass). Gemini-3.1-Pro has low overall Turn Pass (23.6\%) but comparatively high APR after a passing
% state on presentation (70.3\%) and interaction tasks (79.6\%). High APR does not necessarily imply high Turn Pass:
% a model may preserve a correct artifact once reached while rarely reaching one in the first place.

\paragraph{Finding 2: Tool-grounded tasks remain the most difficult after conditioning on a previous pass.}

Model-averaged TP decreases from 55.5\% on Presentation and 47.6\% on Interaction tasks to 25.0\% on Tool-grounded tasks; model-averaged TP@5 similarly falls to 5.0\% (Figure~\ref{fig:main-result-views}a). Tool-grounded tasks also have denser validation contracts, averaging 11.9 requirements per turn, compared with 3.0 for Presentation and 5.9 for Interaction (Table~\ref{tab:benchmark-overview}). Cross-suite differences therefore reflect the full task design, including both external-state grounding and validation density.

APR exhibits the same suite ordering after conditioning on a passing previous turn: 71.1\% for Presentation, 68.7\% for Interaction, and 52.4\% for Tool-grounded tasks (Table~\ref{tab:main-results}). A fixed-seed post hoc audit identifies 110 failed APR transitions for which sufficient attribution evidence is available. Among these transitions, 58 (52.7\%) involve regression on prior behavior, sometimes together with failure on the new requirement; the remaining 52 (47.3\%) preserve prior behavior but fail to satisfy the new requirement (Table~\ref{tab:apr-attribution-audit}).

% Mean Turn Pass decreases from 55.5\% on presentation and 47.6\% on interaction tasks to 25.0\% on tool-grounded tasks; mean TP@5 falls to 5.0\% (Figure~\ref{fig:main-result-views}a). Tool-grounded tasks also carry denser validation contracts (11.9 requirements per turn versus 3.0 and 5.9; Table~\ref{tab:benchmark-overview}), so the cross-suite gap reflects the complete task construction, including external-state grounding and validation load. APR shows the same ordering after conditioning on a passing previous turn: 71.1\%, 68.7\%, and 52.4\%, respectively (Table~\ref{tab:main-results}). A fixed-seed post hoc audit attributes 110 failed APR transitions with sufficient evidence. Of these, 58 (52.7\%) contain a regression of prior behavior, sometimes alongside a missed new requirement; 52 (47.3\%) preserve prior behavior but fail the new requirement (Table~\ref{tab:apr-attribution-audit}).

\paragraph{Finding 3: Reliability declines after the second turn.}
Aggregate pass rates are similar at the first two turn positions but fall to 39.4\% at turn 3 and 35.1\% at turn 4 (Figure~\ref{fig:main-result-views}b). The largest aggregate decline occurs between turns 2 and 3. This decline is especially pronounced for Tool-grounded tasks, whose pass rate falls from 39.5\% at turn 2 to 18.5\% at turn 3 and reaches 14.0\% at turn 4. These results indicate that difficulty increases once models must incorporate multiple revisions while maintaining prior interface behavior and state consistency.

% Aggregate pass rates are similar on the first two turns, then fall to 39.4\% on turn 3 and 35.1\% on turn 4
% (Figure~\ref{fig:main-result-views}b). The largest aggregate drop occurs between turns 2 and 3, particularly for
% tool-grounded tasks, where pass rate falls from 39.5\% to 18.5\% and reaches 14.0\% on turn 4.

\subsection{Robustness and Evaluator Diagnostics}
\label{sec:robustness-diagnostics}

The main experiment uses one generation per model--task--turn slot and MiMo-V2.5 as both the interaction actor and evaluator. We therefore conduct additional studies of actor--evaluator sensitivity, evaluator evidence, and regeneration stability. Appendix~\ref{app:diagnostic-resource-tables} provides the complete protocols and results.

\paragraph{Actor--evaluator sensitivity.}
Across 120 stratified cases, agreement with human annotations ranges from 86.1\% to 91.4\% across the tested actor--evaluator pairings. The default MiMo--MiMo configuration achieves 89.7\% agreement (Table~\ref{tab:actor_evaluator_sensitivity}), indicating that its performance is within the range of the alternative configurations.

\paragraph{Evaluator evidence ablation.}
On a separate balanced reference set of 240 cases, the full-evidence evaluator achieves 87.5\% accuracy against majority-vote human labels. Removing the actor trace produces the largest decrease, reducing accuracy to 55.0\%. Accuracy decreases to 63.8\% without the private evaluation reference, 67.5\% without the source code, and 78.3\% without the screenshot (Table~\ref{tab:evaluator-evidence-ablation}). These results indicate that no single artifact view is sufficient and that interaction traces are particularly important for identifying behavioral failures. Appendix~\ref{app:evaluator-evidence-ablation} provides the complete ablation protocol.

\begin{table}[t]
\centering
\scriptsize
\setlength{\tabcolsep}{2.4pt}
\renewcommand{\arraystretch}{1.04}
\begin{tabular}{@{}lrrrrr@{}}
\toprule
\textbf{Configuration} & \textbf{Acc.} & \textbf{$\Delta$ Acc.} & \textbf{Prec.} & \textbf{Rec.} & \textbf{F1} \\
\midrule
\textbf{Full evidence}
& \textbf{87.5}
& --
& \textbf{86.9}
& \textbf{88.3}
& \textbf{87.6} \\
\midrule
No source code
& 67.5 & $-20.0$ & 67.2 & 68.3 & 67.8 \\
No screenshot
& 78.3 & $-9.2$ & 78.3 & 78.3 & 78.3 \\
No actor trace
& 55.0 & $-32.5$ & 54.7 & 58.3 & 56.5 \\
No private reference
& 63.8 & $-23.7$ & 92.3 & 30.0 & 45.3 \\
\bottomrule
\end{tabular}
\caption{\textbf{Evaluator evidence ablation.} Decision accuracy on a fixed set of 240 balanced cases (120 passing, 120 failing; passing is positive). $\Delta$ Acc. is the percentage-point change from full evidence. Each row removes one source.}
\label{tab:evaluator-evidence-ablation}
\end{table}

\begin{figure}[t]
\centering
\includegraphics[width=0.9\columnwidth]{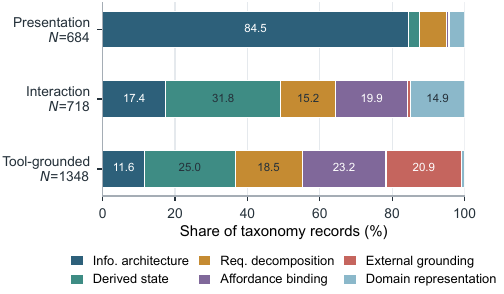}
\caption{\textbf{Diagnostic failure mechanisms.} Values are percentages over all executed non-passing calls in each suite; $N$ denotes the number of calls. Cross-turn preservation is measured separately by APR and TP@5.}
\label{fig:failure-taxonomy-diagnostic}
\end{figure}

\paragraph{Regeneration stability.}
Table~\ref{tab:qwen_regeneration_stability} reports three independent runs of Qwen3.6-Plus. Overall TP ranges from 39.7\% to 46.4\%, while the ordering Presentation (>) Interaction (>) Tool-grounded remains consistent across all three runs. Thus, absolute performance varies across generations, but the principal suite-level difficulty pattern is stable.

% Table~\ref{tab:qwen_regeneration_stability} reports three independent Qwen3.6-Plus runs. Turn Pass spans
% 39.7\%--46.4\%, with Presentation $>$ Interaction $>$ Tool-grounded throughout.

\begin{figure*}[t]
\centering
\includegraphics[width=0.94\textwidth]{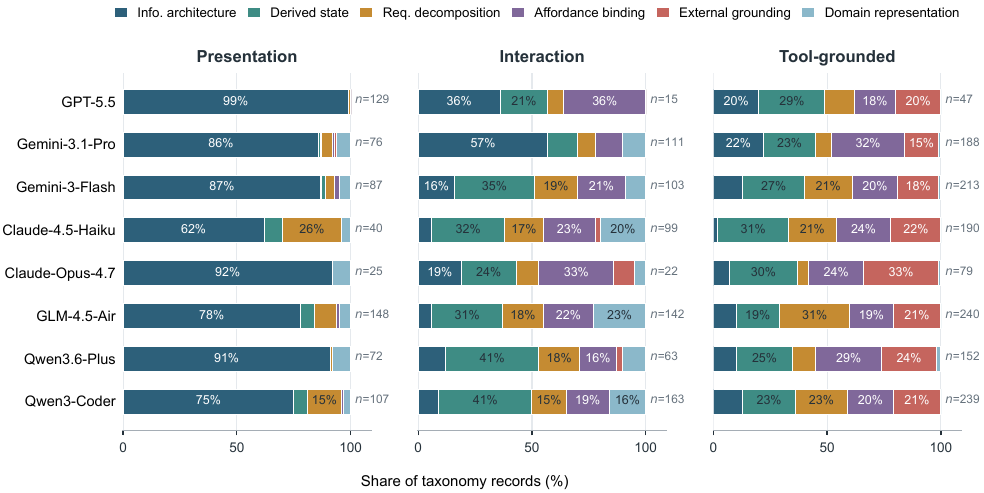}
\caption{\textbf{Failure composition by model and suite.} Each bar is the within-row distribution over six diagnostic mechanisms; $n$ gives the number of executed non-passing calls in that row.}
\label{fig:failure-taxonomy-by-model}
\end{figure*}

\FloatBarrier
\flushbottom

\section{Analysis and Discussion}
% Keep the two wide diagnostic figures together when they fit; restore the
% class defaults after the section so later floats are unaffected.
\setcounter{dbltopnumber}{2}
\renewcommand{\dbltopfraction}{0.95}
\begin{figure*}[t]
\centering
\begin{minipage}[t]{0.437\textwidth}
\centering
\begin{minipage}[c][0.19\textheight][c]{\linewidth}
\centering
\includegraphics[trim=0 0 208bp 0,clip,height=0.19\textheight]{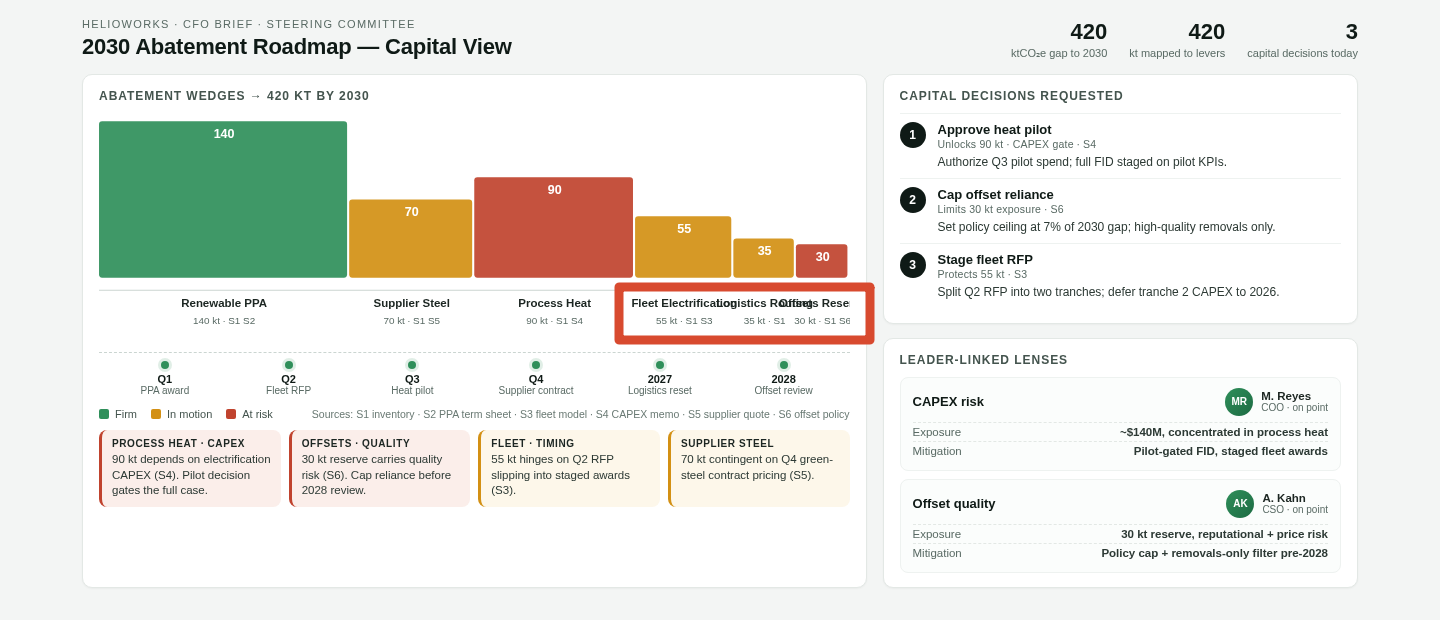}
\end{minipage}\par\smallskip
\small\textbf{(a) Presentation}
\end{minipage}\hfill
\begin{minipage}[t]{0.17\textwidth}
\centering
\begin{minipage}[c][0.19\textheight][c]{\linewidth}
\centering
\includegraphics[height=0.19\textheight]{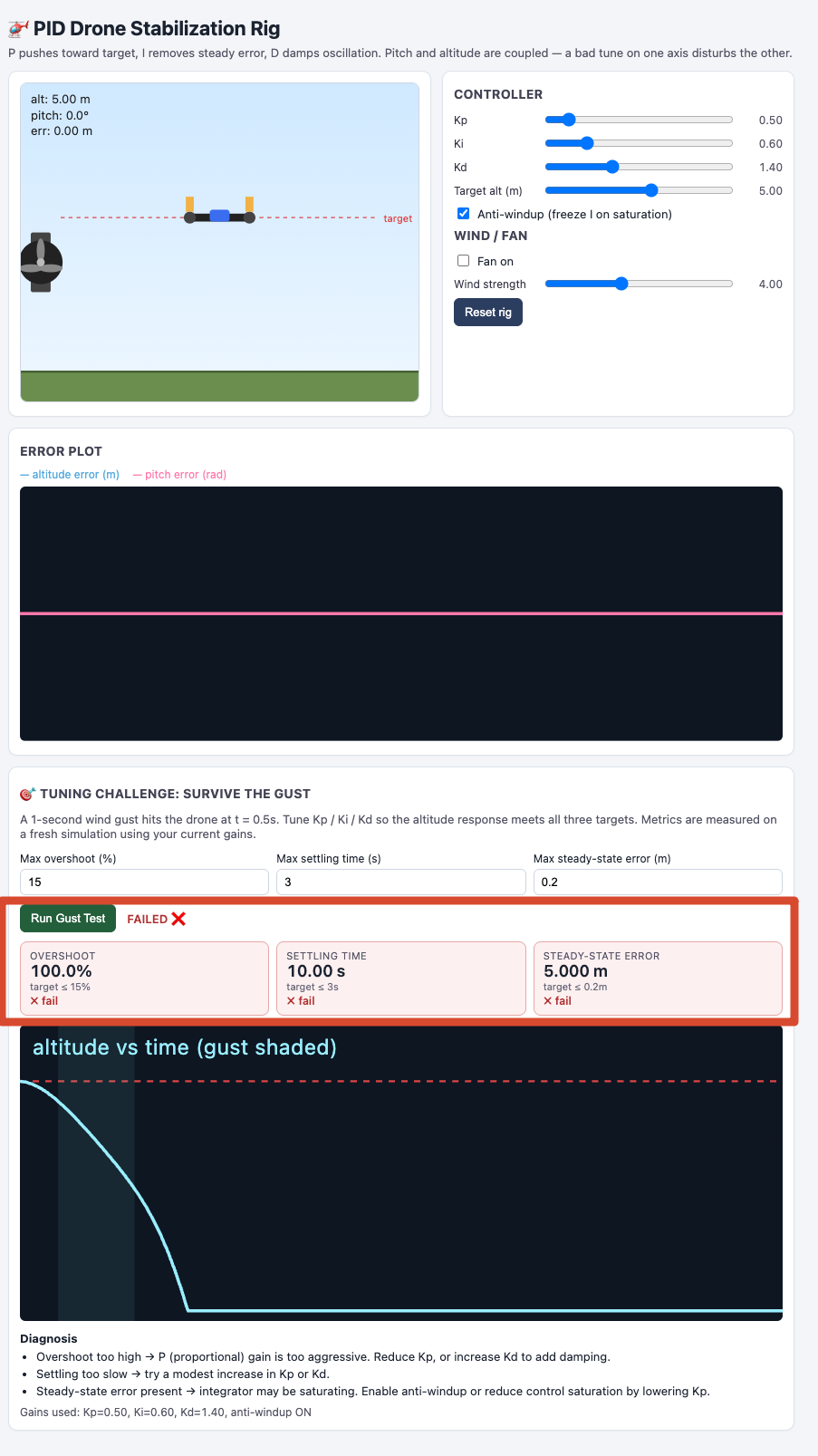}
\end{minipage}\par\smallskip
\small\textbf{(b) Interaction}
\end{minipage}\hfill
\begin{minipage}[t]{0.385\textwidth}
\centering
\begin{minipage}[c][0.19\textheight][c]{\linewidth}
\centering
\includegraphics[height=0.19\textheight]{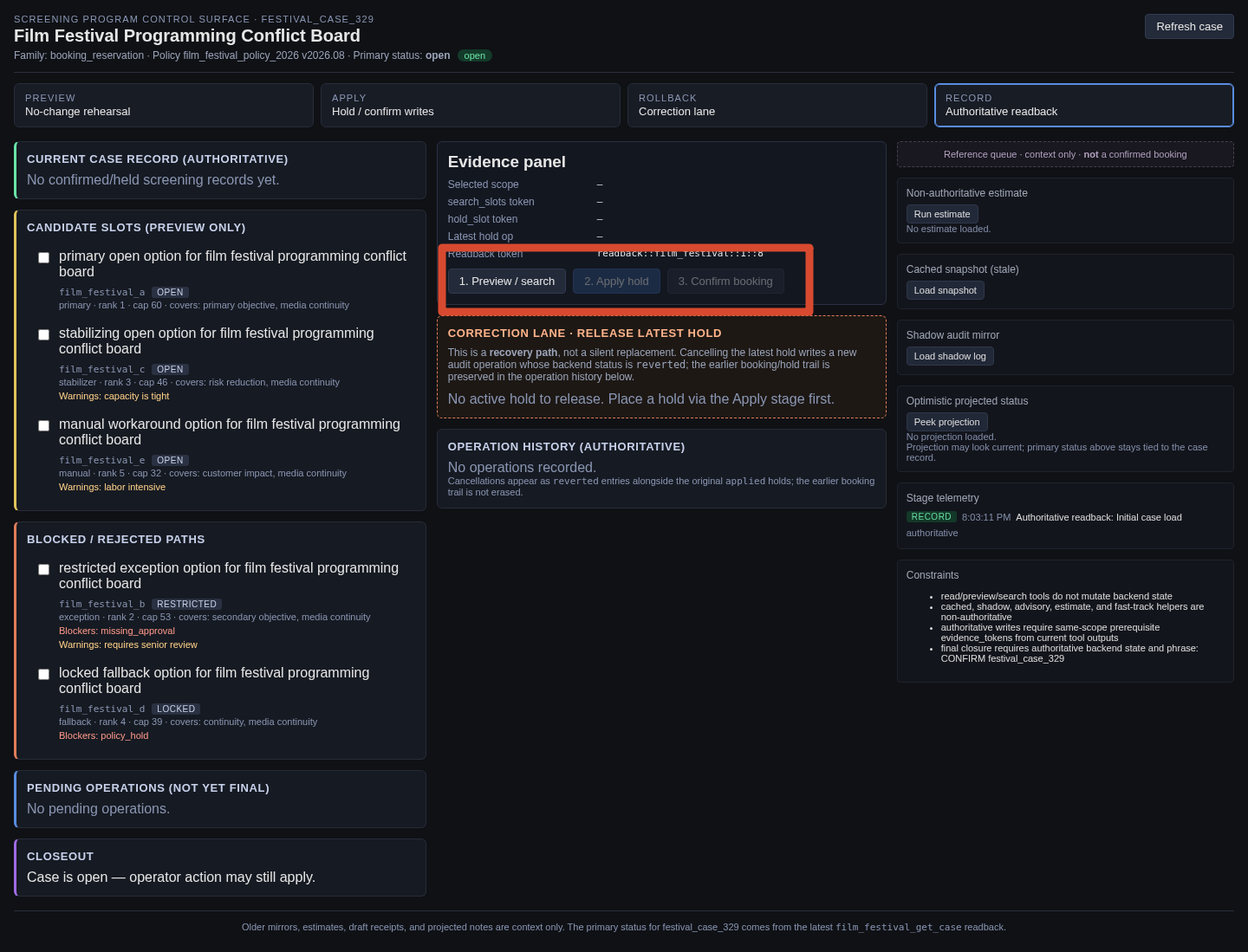}
\end{minipage}\par\smallskip
\small\textbf{(c) Tool-grounded}
\end{minipage}
\caption{\textbf{Three diagnostic failures.} Red boxes mark (a) merged chart labels, (b) metrics that remain stale after gain changes, and (c) controls blocked by mismatched backend evidence tokens.}
\label{fig:case-study-overview}
\end{figure*}

\subsection{Diagnostic Failure Taxonomy}
\label{sec:failure-taxonomy}

We classify each failed execution into one of six interface-maintenance mechanisms: \emph{information architecture}, for poorly organized or unreadable content; \emph{domain representation}, for incorrect task-specific abstractions; \emph{requirement decomposition}, for omitted constraints or workflow steps; \emph{affordance binding}, for visible but unwired controls; \emph{derived-state propagation}, for dependent views that remain stale after local state changes; and \emph{external-state grounding}, for contradictions between the interface and tool or runtime state. The analysis covers all 2,750 executed non-passing calls, each assigned to one primary mechanism: 859 information-architecture, 586 derived-state-propagation, 460 affordance-binding, 410 requirement-decomposition, 289 external-state-grounding, and 146 domain-representation failures. Episode-level reliability and cross-turn retention are analyzed using TP@5 and APR, respectively.

Before large-scale labeling, we constructed and froze the six-label codebook using the evaluation rubrics, private validation criteria, and representative failures. On a category-balanced sample of 120 classified failures (20 per category), a separate diagnostic judge achieves 90.3\% mean exact-label agreement with three blinded non-author annotators. Mean multiclass Cohen's $\kappa$ between the judge and individual annotators is 0.89, and Fleiss' $\kappa$ among the human annotators is 0.91. Appendix~\ref{app:failure-taxonomy-validation} provides the complete labeling and validation protocol.

% We built and froze the six-label codebook from evaluation rubrics, private validation criteria, and representative failures before large-scale labeling. On a category-balanced sample of 120 classified failures (20 per category), a separate diagnostic judge reaches 90.3\% mean exact-label agreement with three blinded non-author annotators (mean multiclass Cohen's $\kappa=0.89$; human Fleiss' $\kappa=0.91$). Appendix~\ref{app:failure-taxonomy-validation} gives the full labeling protocol.

Figure~\ref{fig:failure-taxonomy-diagnostic} shows that the dominant bottleneck varies by scenario. Presentation failures primarily involve \emph{information architecture}: requested facts lack a clear hierarchy, overlap visually, or appear clipped. Interaction failures more often involve \emph{derived-state propagation} and \emph{affordance binding}, producing stale dependent views or controls that appear functional but are not wired to behavior. Tool-grounded failures span these mechanisms as well as \emph{external-state grounding} and \emph{requirement decomposition}. Reliable tool-grounded \genui{} therefore requires coordinated management of interface and runtime state, not merely valid API invocation. The model-level results in Figure~\ref{fig:failure-taxonomy-by-model} show that these suite-level patterns recur across model families rather than being driven by a single model. Section~\ref{sec:case-studies} traces three mechanisms through representative artifacts.

% Figure~\ref{fig:failure-taxonomy-diagnostic} shows that the dominant bottleneck shifts across scenarios. Presentation failures mainly involve \emph{information architecture}: requested facts have weak hierarchy, overlap, or appear clipped. Interaction failures more often involve \emph{derived-state propagation} and \emph{affordance binding}, with stale dependent views or unwired controls. Tool-grounded failures are distributed across these two labels, \emph{external-state grounding}, and \emph{requirement decomposition}. Tool-grounded \genui{} therefore depends on coupled UI--runtime state management, not just valid API calls. The model-level view in Figure~\ref{fig:failure-taxonomy-by-model} shows that the suite patterns recur across model families rather than being driven by one model. Section~\ref{sec:case-studies} traces three mechanisms through representative artifacts.

\subsection{Case Studies}
\label{sec:case-studies}

We examine three executed turns from Claude-Opus-4.7, the strongest model overall, to reveal how a plausible interface can conceal a broken mechanism.

\paragraph{Presentation: Carbon Abatement Roadmap Board, Turn 2.}
The model preserves all six requested wedges, source markers S1--S6, the timeline, and the decision callouts.
Execution and Alignment both score 5, but Presentation scores 3 because the SVG chart merges several x-axis
labels into one unreadable string. The requested information is present, yet its layout makes it unusable. The screenshot exposes this
\emph{information architecture} failure (Figure~\ref{fig:case-study-overview}a).

\paragraph{Interaction: Drone PID Wind Disturbance Lab, Turn 5.}
The generated challenge panel contains target inputs, a \emph{Run Gust Test} button, pass/fail badges, and diagnosis text, earning Presentation 5 and Alignment 4. The actor changes $K_p$ from $2.20$ to $1.0$, $1.5$, and $0.5$, but every run returns 100.0\% overshoot, 10.00\,s settling, and 5.000\,m SSE. 
Although the displayed gain updates, the dependent simulation metrics remain fixed, resulting in an Execution score of 3. The actor trace and source difference expose this \emph{derived-state propagation} failure (Figure~\ref{fig:case-study-overview}b).

\paragraph{Tool-grounded: Film Festival Programming Conflict Board, Turn 4.}
The interface renders the requested correction lane, action rail, and evidence tokens, earning Presentation 5 and Alignment 4. Its tool-calling code, however, sends stale or mismatched prerequisite tokens to the backend, causing every \texttt{hold\_slot} call to return \texttt{missing\_prerequisite\_token\_slots}. After 17 actor steps across multiple interaction paths, no hold succeeds and the correction lane remains unreachable. Execution therefore scores 3. The runtime logs reveal the mismatch between interface state and backend state, identifying an \emph{external-state grounding} failure (Figure~\ref{fig:case-study-overview}c).

These cases illustrate different failures require different evidence: the presentation defect is visible in the screenshot, the stale derived state emerges from the actor trace and source difference, and the tool-state mismatch appears in runtime logs.

% The interface renders the requested correction lane, action rail, and evidence tokens, earning Presentation 5 and
% Alignment 4. Its tool-calling code sends stale or mismatched tokens to the backend, however, so every
% \texttt{hold\_slot} call returns \texttt{missing\_prerequisite\_token:search\_slots}. After 17 actor steps across
% several interaction paths, no hold succeeds and the correction lane remains unreachable. Execution scores 3;
% runtime logs identify the defect as \emph{external-state grounding} (Figure~\ref{fig:case-study-overview}c). Each
% failure is exposed by a different source: the presentation error by the screenshot, stale derived state by the actor
% trace and source diff, and the tool failure by runtime logs.

\subsection{Implications for Generative UI Evaluation}

The results distinguish two aspects of reliability: whether a model reaches a correct artifact and whether it preserves correctness across subsequent revisions. They also show that reliable diagnosis requires matching evidence to mechanism. Screenshots reveal visual and organizational defects; actor traces and source changes reveal broken interactions and stale derived state; and runtime logs reveal inconsistencies between the interface and external systems. No single score or evidence surface captures all three. Generative UI evaluation should therefore combine outcome-level reliability metrics with mechanism-level diagnosis.
%\FloatBarrier
\setcounter{dbltopnumber}{2}
\renewcommand{\dbltopfraction}{0.7}

\section{Conclusion}
We introduced \benchmark{}, a benchmark that evaluates generative UI as multi-turn maintenance of an executable artifact. By combining browser execution with visual, behavioral, source-level, and runtime evidence, it measures both turn-level correctness and reliability across revisions. Experiments across model families show that strong single-turn performance does not reliably translate into sustained episode success, particularly for stateful and tool-grounded interfaces.

Our diagnostic analysis further shows that failure mechanisms vary by task, spanning information architecture, interaction wiring, derived-state propagation, requirement decomposition, and external-state grounding. These findings demonstrate that neither a final screenshot nor an isolated turn score adequately captures generative UI reliability. Evaluation must instead assess whether interface behavior, dependent state, external state, and assistant claims remain synchronized as the artifact evolves.

% \benchmark{} tests whether a model can revise the same executable web interface across five turns, using 150 tasks
% and 750 turns. Browser execution, screenshots, actor traces, source and DOM evidence, and runtime logs support
% turn-, episode-, and transition-level measures. Across eight models, the strongest model reaches 74.9\% Turn Pass
% but only 37.3\% TP@5, while tool-grounded APR is 52.4\%. A blinded human study finds that the automatic
% evaluator agrees with majority-vote human labels in 86.7\% of cases.

% Failure mechanisms shift with the task: presentation failures center on information architecture, interaction
% failures on derived-state propagation and affordance binding, and tool-grounded failures additionally expose
% external-state grounding and requirement decomposition. A final screenshot or a single-turn score misses these
% breakdowns. Generative UI evaluation should therefore test whether interface behavior, derived state, external
% state, and assistant text remain synchronized as the artifact evolves.

\section*{Limitations}
\benchmark{} uses human-authored five-turn episodes in a fixed React/Vite browser environment. Its scope excludes naturally occurring interaction logs, other UI frameworks, accessibility requirements, and device conditions.

APR is an outcome-level retention measure rather than a causal attribution metric. A failed transition may reflect regression on prior behavior, failure to satisfy the new request, or both. Our post hoc audit distinguishes prior-regression-containing from new-requirement-only failures on a sample, but comprehensive attribution would require replaying all prior-turn validation contracts against each later artifact.

Tool-grounded tasks use deterministic mock runtimes to ensure reproducibility. They therefore evaluate controlled interface--runtime synchronization but do not capture operational challenges of live services, including latency, authentication and permission failures, outages, rate limits, and API changes.

\section*{Ethical Considerations}
\benchmark{} uses synthetic tasks, mock external state, and deterministic tool environments rather than private user data or live third-party services. This reduces privacy risk and supports reproducible evaluation. Before release, we manually screened task prompts, mock fixtures, and public resources for personally identifying information and offensive content. The released artifacts preserve the distinction between generator-visible inputs and private validation material.

Generated interfaces can mislead users when they expose polished but unwired controls, summaries, confirmations, or status labels that are not grounded in executable behavior. \benchmark{} probes these risks through evidence-grounded execution, interaction, and tool-state checks. However, it remains a diagnostic benchmark in a controlled setting and does not by itself establish the safety of generated interfaces in deployment.

\section*{Acknowledgments}

Generative AI tools were used to assist with language polishing and
manuscript editing. All generated suggestions were reviewed and verified
by the authors, who take full responsibility for the content of the paper.

\bibliography{references}

\appendix
\raggedbottom
% Appendix pages contain several single-column figures and tables. Allow two
% bottom floats on a page so a short case-study figure is not deferred into the
% prompt appendix and does not create an otherwise avoidable final page.
\setcounter{topnumber}{3}
\setcounter{bottomnumber}{2}
\setcounter{totalnumber}{5}
\renewcommand{\topfraction}{0.9}
\renewcommand{\bottomfraction}{0.5}
\renewcommand{\textfraction}{0.08}
\renewcommand{\floatpagefraction}{0.75}

\section{Benchmark Artifacts and Release}

\subsection{Benchmark Card}
\label{app:benchmark-card}
\begin{itemize}
    \item \textbf{Intended use:} comparing models on multi-turn generative UI construction and revision.
    \item \textbf{Not intended for:} certifying safety or production readiness of generated applications.
    \item \textbf{Primary metrics:} Turn Pass, TP@5, APR, CPT, and dimension scores.
    \item \textbf{Required artifacts:} generated source files, assistant text, build logs, browser snapshots, interaction traces, tool logs when applicable, evaluator outputs, and retention summaries.
    \item \textbf{Release:} Version \mbox{\texttt{v1.0}} is available under the MIT License at
    \url{https://github.com/MAPS-research/EvoGenUI-Bench}; Appendix~\ref{app:artifact-release} gives the release protocol.
\end{itemize}

\subsection{Artifact Release and Licensing}
\label{app:artifact-release}

\paragraph{Repository.}
The complete code, data, and pipeline are available under the MIT License at
\url{https://github.com/MAPS-research/EvoGenUI-Bench}, with a README and reproduction instructions.

\paragraph{Versioning.}
The benchmark version used in this paper is \mbox{\texttt{v1.0}}.
Any reported evaluation result should identify the benchmark version, code commit, task manifest, prompt versions, and evaluator/actor configuration, because changes to tasks, prompts, hidden validation contracts, or runtime fixtures can affect scores.

\paragraph{Licensing.}
The MIT License covers the code, task data, mock fixtures, public resources, documentation, and benchmark artifacts. Model outputs and interaction traces use the same license where model-provider terms permit it; covered materials are marked in the repository.

\paragraph{Public and evaluator-only components.}
Generator-visible materials include user turns, public context, public tool/resource contracts, prompt templates,
runtime code, and evaluation scripts. Evaluator-only materials include hidden validation contracts, actor-only
facts, backend fixtures, runtime state, and evaluator-specific evidence requirements. Both are released for
auditing and reproduction, but the execution pipeline never exposes evaluator-only fields to generator models.

\section{Task Suite Design}

\subsection{Task Authoring Workflow}
\label{app:task-construction}

The authoring workflow is:

\begin{enumerate}
    \item Choose a domain and interface artifact type, such as a planning dashboard, evidence board, simulator,
    reasoning workbench, map, or action console. The domain is checked against the suite manifest to avoid
    duplicate domain labels and near-duplicate interface types.
    \item Write a five-turn user trajectory in which each turn introduces a cumulative revision to the same
    artifact rather than an independent mini-app.
    \item Specify private validation requirements for each turn, including requirements from earlier turns that
    remain active.
    \item Bind each requirement to one or more evidence surfaces: screenshot, DOM, interaction trace, tool log,
    external state, source changes, visible UI text, or assistant text.
    \item Add task metadata describing the targeted challenge, such as information hierarchy, linked
    derived-state propagation, state-machine invariants, affordance binding, tool read/write grounding,
    authoritative readback, or text--artifact calibration.
    \item Cross-validate the task with a different human reviewer, who checks ambiguity, domain duplication,
    rubric leakage, observability of the validation contract, and whether later turns genuinely require
    multi-turn maintenance rather than independent regeneration.
\end{enumerate}

Exact positions or wording are enforced only when the user requests them; otherwise, semantically equivalent
layouts and interaction paths remain valid.

% For example, a tool-grounded pharmacy task asks the model to maintain a prior-authorization appeal console across
% five turns: search and stage records, rehearse without writing, save with readback, revise the saved choice, and
% produce an audit summary. The validation contract binds success to staged-versus-committed state, tool logs,
% post-write readback, and assistant text that does not overclaim completion.

\subsection{Common Task Schema}
\label{app:task-json-format}

Each task in \benchmark{} is a five-turn JSON episode. The abridged structure below retains the fields that define
generator-visible input and private validation; curation metadata is omitted.

\begin{jsonlistingbox}{Shared Task Structure (abridged)}
{
  "task_id": "...",
  "domain": "...",
  "suite": "presentation_ui | interactive_tool_ui | tool_grounded_action_ui",
  "core_interaction": ["..."],

  "metadata": {
    "task_world": {
      "public_context": "..."
    },
    "validation_contract": {
      "validation_scenarios": [
        {
          "turn": 1,
          "evidence_requirements": [
            {
              "surface": "screenshot / DOM / actor trace / tool log / runtime state / source / assistant text",
              "expect": "..."
            }
          ]
        }
      ]
    }
  },

  "turns": [
    {
      "turn": 1,
      "prompt": "..."
    }
  ]
}
\end{jsonlistingbox}

The generator receives the public task context and turn prompts. The validation contract remains private and binds
expected behavior to one or more observable sources. \texttt{core\_interaction} records the mechanisms stressed by
the task for suite-level analysis.

Tool-grounded tasks add public tool contracts plus private bindings and backend state.

\begin{jsonlistingbox}{Tool-Grounded Extension (abridged)}
{
  "tools": [
    {
      "name": "...",
      "mode": "read | write",
      "mock_contract": {
        "fixture_id": "...",
        "handler_ref": "...",
        "semantic_contract": {}
      }
    }
  ],

  "initial_state": {}
}
\end{jsonlistingbox}

The model sees each tool's name, description, and public schema. Fixture bindings, initial state, and validation
requirements remain private to execution and evaluation.

\subsection{Representative Task Example}
\label{app:example-task}

\emph{Pharmacy Prior Authorization Control Surface} asks the model to maintain an appeal console for case
\texttt{rx\_case\_592}. Its five revisions progress from staging and no-write rehearsal to guarded mutation,
readback, recovery, and audit closeout. The table-and-detail workspace separates current state, candidate actions,
blocked alternatives, rehearsal output, pending work, and final readiness. The trajectory tests grounded read/write
behavior, resistance to non-authoritative helpers, and audit-history preservation.

\begin{table*}[ht]
\centering
\small
\begin{tabularx}{\linewidth}{lX}
\toprule
\textbf{Tool group} & \textbf{Examples} \\
\midrule
Authoritative read tools & \texttt{get\_case}, \texttt{search\_records}, \texttt{get\_record\_details}, \texttt{compare\_records}, \texttt{get\_audit\_log} \\
Authoritative write tool & \texttt{save\_selection} \\
Decoy or non-authoritative helpers & cached snapshot, estimate, advisory check, fast-track request, shadow audit log, optimistic status, stale evidence token, draft receipt, commit-phrase validation \\
Near-miss same-domain tools & entity resolution, profile lookup, capacity pools, simulated allocation, slot search, policy note \\
\bottomrule
\end{tabularx}
\caption{\textbf{Tool groups.} Authoritative, decoy, and near-miss tools in the pharmacy task.}
\label{tab:example-tool-groups}
\end{table*}

The hidden contract checks observable evidence rather than a reference implementation: family-tool use, backend-derived selected and blocked records, post-write readback, a visible interaction path, and assistant text consistent with backend state. Later turns add no-write rehearsal, scoped writes, reversible recovery, and evidence-chain reconstruction.

\section{Evaluation Protocol and Metrics}

\subsection{Evaluation Rubrics}
\label{app:evaluation-rubrics}

The rubric expands the three evaluator-scored dimensions in Section~\ref{sec:evaluation-dimensions}. Each uses a
1--5 scale and passes at scores of 4 or 5. Appendix~\ref{app:evaluation-metrics} defines the statistical metrics
derived from these turn-level decisions.

\paragraph{Presentation.}
Presentation assesses the rendered UI quality: layout, hierarchy, readability, styling, discoverability, and whether screenshot evidence shows a coherent domain-appropriate interface rather than a sparse scaffold, wireframe, or visibly broken page. For screenshot-grounded artifacts, DOM labels and source code can show that objects exist, but they cannot prove that labels are readable, unclipped, or visually coherent.

\paragraph{Execution.}
Execution assesses whether the current turn's requested requirements are implemented as working UI behavior or visible, evidence-backed state. It covers required surfaces, interaction wiring, state updates, tool/resource calls, derived outputs, committed readback, confirmations, and relevant still-valid prior requirements. A button, form, raw tool response, or optimistic confirmation does not pass if the requested result or dependent surface is missing, stale, disconnected, or unsupported by runtime evidence.

\paragraph{Alignment.}
Alignment assesses whether assistant text, generated source, visible UI, interaction observations, and runtime logs describe the same concrete behavior. It penalizes unsupported claims, stale descriptions, claims of unavailable capabilities, claimed saved/submitted/filtered/computed state without evidence, and mismatches between the UI state and the assistant's explanation.

\subsection{Evaluation Metrics}
\label{app:evaluation-metrics}

Section~\ref{sec:metrics} defines the official turn-, episode-, and transition-level metrics. Countable-only
variants are diagnostic rather than official: they remove provider failures for which no model response was
obtained. Interaction-agent status and static code analysis are retained for evidence and attribution rather than
used as direct pass gates.

Task-level bootstrap 95\% confidence intervals resample complete episodes with replacement, preserving
within-episode correlations. Table~\ref{tab:metric-uncertainty} reports percentile intervals and the
TP@5 independence baselines.

\subsubsection{Post Hoc Attribution of Failed APR Transitions}
\label{app:apr-attribution-audit}

We audit a fixed-seed sample of eligible transitions in which the previous turn passed and the following turn failed. For
each transition, annotators inspect the previous user request, previous passing evidence, current user request,
current failure evidence, generated source and DOM evidence, screenshots, browser interaction traces, and
tool/runtime logs when available. The audit is diagnostic only and does not affect the official APR calculation.

We assign each attributable failed transition to one of two categories. \emph{Prior-regression-containing} means
that at least one behavior, state, surface, or still-valid requirement supported by the previous passing artifact
was removed, contradicted, disconnected, or made stale in the next revision. The same transition may also contain a
failure on the newly introduced requirement. \emph{New-requirement-only} means that the previous behavior remains
largely intact and the failed transition is primarily explained by the current turn's newly introduced or revised
requirement.

\begin{table}[ht]
\centering
\small
\begin{tabular}{lrr}
\toprule
\textbf{Attribution category} & \textbf{Count} & \textbf{Share} \\
\midrule
Prior-regression-containing & 58 & 52.7\% \\
New-requirement-only & 52 & 47.3\% \\
\bottomrule
\end{tabular}
\caption{\textbf{Failed APR transition audit.} Post hoc attribution of $N=110$ failed APR transitions with sufficient evidence.}
\label{tab:apr-attribution-audit}
\end{table}

\section{Experimental Configuration and Validation}
% Queue the first two full-width tables before the subsection prose so they can
% share the next page top instead of leaving a half-empty text column.
\setcounter{dbltopnumber}{4}
\setcounter{totalnumber}{5}
\renewcommand{\dbltopfraction}{0.95}
\renewcommand{\textfraction}{0.05}
\makeatletter
\setlength{\@dblfptop}{0pt}
\setlength{\@dblfpsep}{5pt plus 1pt minus 1pt}
\setlength{\@dblfpbot}{0pt plus 1fil}
\makeatother

\begin{table*}[t]
\centering
\captionsetup{skip=2pt}
\scriptsize
\setlength{\tabcolsep}{5pt}
\renewcommand{\arraystretch}{1.08}
\begin{tabular}{@{}llcc@{}}
\toprule
\textbf{Model Alias} & \textbf{Exact Model Identifier} & \textbf{Temp.} & \textbf{Output Token Cap} \\
\midrule
GPT-5.5 & gpt-5.5 & 0 & 32,768 \\
Gemini-3.1-Pro & gemini-3.1-pro-preview & 0 & 32,768 \\
Gemini-3-Flash & gemini-3-flash-preview & 0 & 32,768 \\
Claude-4.5-Haiku & claude-haiku-4-5-20251001 & 0 & 32,768 \\
Claude-Opus-4.7 & claude-opus-4-7 & 0 & 32,768 \\
GLM-4.5-Air & glm-4.5-air & 0 & 32,768 \\
Qwen3.6-Plus & qwen3.6-plus & 0 & 32,768 \\
Qwen3-Coder & qwen3-coder-480b-a35b-instruct & 0 & 32,768 \\
\bottomrule
\end{tabular}

\vspace{4pt}
\textbf{Requested-slot accounting}
\par\smallskip
\setlength{\tabcolsep}{4pt}
\begin{tabular}{@{}lrr@{}}
\toprule
\textbf{Slot category} & \textbf{Count} & \textbf{Share} \\
\midrule
Executed generation calls & 5,310 & 88.5\% \\
Blocked after earlier build failure & 443 & 7.4\% \\
Blocked after earlier invalid output & 247 & 4.1\% \\
\midrule
\textbf{Total requested slots} & \textbf{6,000} & \textbf{100.0\%} \\
\bottomrule
\end{tabular}
\caption{\textbf{Evaluated models and requested-slot accounting.} The upper table lists generator identifiers and
inference settings; output token cap is the configured completion limit, not observed mean length. The lower table
separates executed calls from unexecuted downstream slots. Unexecuted slots remain failures for Turn Pass,
TP@5, and CPT.}
\label{tab:evaluated_models}
\label{tab:slot-accounting}
\end{table*}

\begin{table*}[t]
\centering
\captionsetup{skip=2pt}
\scriptsize
\setlength{\tabcolsep}{4pt}
\begin{tabular}{lllcccc}
\toprule
\textbf{Check} & \textbf{Actor} & \textbf{Evaluator}
& \textbf{Interaction} & \textbf{Presentation} & \textbf{Tool-grounded}
& \textbf{Overall} \\
\midrule
Default & MiMo & MiMo
& 90.8\% & 88.3\% & 90.0\% & 89.7\% \\
\midrule
Fixed actor & MiMo & Qwen3.6-Plus
& 88.3\% & 85.8\% & 87.5\% & 87.2\% \\
Fixed actor & MiMo & GPT-5.5
& \textbf{91.7\%} & 89.2\% & \textbf{90.8\%} & \textbf{90.6\%} \\
\midrule
Fixed evaluator & Qwen3.6-Plus & MiMo
& 89.2\% & 86.7\% & 88.3\% & 88.1\% \\
Fixed evaluator & GPT-5.5 & MiMo
& \textbf{91.7\%} & 88.3\% & 90.0\% & \textbf{90.0\%} \\
\midrule
Same model & Qwen3.6-Plus & Qwen3.6-Plus
& 87.5\% & 84.2\% & 86.7\% & 86.1\% \\
Same model & GPT-5.5 & GPT-5.5
& \textbf{92.5\%} & \textbf{90.0\%} & \textbf{91.7\%} & \textbf{91.4\%} \\
\bottomrule
\end{tabular}
\caption{\textbf{Actor--evaluator sensitivity.} Diagnostics on $N=120$ cases (40 per suite). MiMo denotes MiMo-V2.5.
Each case receives three human annotations; suite and overall percentages therefore aggregate 120 and 360
individual annotations per configuration, respectively. Human Agreement is the percentage of those annotations
that support the automatic decision. Bold entries exceed the default MiMo--MiMo configuration within the same
suite.}
\label{tab:actor_evaluator_sensitivity}
\end{table*}

\subsection{Evaluated Models}
\label{app:evaluated-models}

Table~\ref{tab:evaluated_models} lists the generator models and inference settings used for the final reported evaluation slice. Once the final run set was selected, completed model outputs were frozen for downstream analysis.

All settings refer to generator calls; the token value is a completion cap, not mean output length. The default MiMo-V2.5 actor and evaluator use temperature 0, output caps of 4,096 and 8,192 tokens, respectively, and fixed prompts across generator models.

\subsection{Robustness Diagnostics}
\label{app:diagnostic-resource-tables}

% Queue the remaining diagnostic tables before the subsection prose.
\begin{table*}[t]
\centering
\captionsetup{skip=2pt}
\scriptsize
\setlength{\tabcolsep}{4pt}
\begin{tabular}{llccccc}
\toprule
\textbf{Suite} & \textbf{Run}
& \textbf{Turn Pass}
& \textbf{Pres.}
& \textbf{Exec.}
& \textbf{Align.}
& \textbf{Avg. Score} \\
\midrule
\multirow{4}{*}{Presentation}
& R1 & 60.0\% & 3.86 & 3.71 & 3.82 & 3.80 \\
& R2 & 55.2\% & 3.74 & 3.58 & 3.69 & 3.67 \\
& R3 & 62.0\% & 3.93 & 3.78 & 3.88 & 3.86 \\
& \textbf{Mean $\pm$ SD} & 59.1 $\pm$ 3.5 & 3.84 $\pm$ 0.10 & 3.69 $\pm$ 0.10 & 3.80 $\pm$ 0.10 & 3.78 $\pm$ 0.10 \\
\midrule
\multirow{4}{*}{Interaction}
& R1 & 46.8\% & 3.62 & 3.28 & 3.55 & 3.48 \\
& R2 & 42.8\% & 3.48 & 3.12 & 3.41 & 3.34 \\
& R3 & 49.2\% & 3.70 & 3.37 & 3.63 & 3.57 \\
& \textbf{Mean $\pm$ SD} & 46.3 $\pm$ 3.2 & 3.60 $\pm$ 0.11 & 3.26 $\pm$ 0.13 & 3.53 $\pm$ 0.11 & 3.46 $\pm$ 0.12 \\
\midrule
\multirow{4}{*}{Tool-grounded}
& R1 & 25.2\% & 3.31 & 2.91 & 3.12 & 3.11 \\
& R2 & 21.2\% & 3.17 & 2.72 & 2.95 & 2.95 \\
& R3 & 28.0\% & 3.42 & 3.05 & 3.25 & 3.24 \\
& \textbf{Mean $\pm$ SD} & 24.8 $\pm$ 3.4 & 3.30 $\pm$ 0.13 & 2.89 $\pm$ 0.17 & 3.11 $\pm$ 0.15 & 3.10 $\pm$ 0.15 \\
\midrule
\multirow{4}{*}{Overall}
& R1 & 44.0\% & 3.60 & 3.30 & 3.50 & 3.46 \\
& R2 & 39.7\% & 3.46 & 3.14 & 3.35 & 3.32 \\
& R3 & 46.4\% & 3.68 & 3.40 & 3.59 & 3.56 \\
& \textbf{Mean $\pm$ SD} & \textbf{43.4 $\pm$ 3.4} & \textbf{3.58 $\pm$ 0.11} & \textbf{3.28 $\pm$ 0.13} & \textbf{3.48 $\pm$ 0.12} & \textbf{3.45 $\pm$ 0.12} \\
\bottomrule
\end{tabular}
\caption{\textbf{Regeneration stability.} Three independent Qwen3.6-Plus generations on the full 150-task, 750-turn slice. Pres., Exec., and Align. are mean 1--5 evaluator scores; Avg. Score averages the three dimensions.}
\label{tab:qwen_regeneration_stability}
\end{table*}

\begin{table*}[t]
\centering
\captionsetup{skip=2pt}
\scriptsize
\setlength{\tabcolsep}{3.4pt}
\begin{tabular}{@{}lccccc@{}}
\toprule
\textbf{Model}
& \textbf{Turn Pass [95\% CI]}
& \textbf{TP@5 [95\% CI]}
& \textbf{CPT [95\% CI]}
& \textbf{APR [95\% CI]}
& \textbf{Indep.\ TP@5} \\
\midrule
GPT-5.5          & 61.7\% [56.9, 66.1] & 21.3\% [14.7, 28.0] & 2.21 [1.89, 2.51] & 71.0\% [65.7, 75.8] & 8.7\% \\
Qwen3.6-Plus     & 44.0\% [38.5, 48.4] & 7.3\% [3.3, 11.3]   & 1.32 [1.03, 1.57] & 61.7\% [55.4, 67.5] & 1.3\% \\
Gemini-3-Flash   & 41.7\% [36.8, 46.7] & 4.7\% [1.3, 8.0]    & 1.02 [0.78, 1.27] & 57.8\% [51.7, 63.8] & 1.2\% \\
Claude-Opus-4.7  & 74.9\% [70.8, 79.6] & 37.3\% [30.0, 45.3] & 2.81 [2.51, 3.14] & 83.6\% [79.9, 86.9] & 23.4\% \\
\midrule
Qwen3-Coder      & 23.2\% [18.8, 27.9] & 2.0\% [0.0, 4.7] & 0.77 [0.58, 0.97] & 42.2\% [33.6, 50.5] & $<$0.1\% \\
GLM-4.5-Air      & 21.1\% [16.7, 25.2] & 1.3\% [0.0, 3.3] & 0.60 [0.41, 0.77] & 48.5\% [39.7, 56.1] & $<$0.1\% \\
Gemini-3.1-Pro   & 23.6\% [19.1, 28.3] & 6.7\% [3.3, 11.3] & 0.69 [0.49, 0.91] & 68.6\% [60.8, 75.4] & 0.1\% \\
Claude-4.5-Haiku & 51.1\% [45.9, 56.5] & 13.3\% [8.7, 18.7] & 1.61 [1.31, 1.92] & 65.7\% [59.4, 70.9] & 3.2\% \\
\bottomrule
\end{tabular}
\caption{\textbf{Metric uncertainty.} Task-level bootstrap 95\% confidence intervals for the four aggregate metrics, with the
independence baseline for TP@5 (Indep.\ TP@5). TP, TP@5, and APR are percentages; CPT
is in turns out of five. All rows use task-level episode resampling under the denominator conventions in
Appendix~\ref{app:evaluation-metrics}.}
\label{tab:metric-uncertainty}
\end{table*}

\subsubsection{Evaluator Evidence and Component Ablation}
\label{app:evaluator-evidence-ablation}

The ablation uses a fixed reference set of 240 frozen artifact packets, balanced between 120 passing and 120
failing cases. Each packet includes the user request, generated UI, assistant response, build result, browser trace,
final DOM, screenshot, and runtime logs.

We rerun the evaluator on each frozen artifact with full evidence or with one source removed: generated code,
screenshot, actor trace, or private evaluation reference. Removing the actor trace leaves the final DOM,
screenshot, and runtime logs; removing the private reference leaves public task context and observable artifacts.
Metrics compare the automatic turn-level decision with the fixed reference label, treating passing turns as the
positive class.

\subsubsection{Failure Taxonomy Construction and Validation}
\label{app:failure-taxonomy-validation}

Before large-scale labeling, we grouped recurring errors from the evaluation rubrics, private validation criteria,
and manually inspected failures into six categories, merged overlaps, and froze the codebook.

A diagnostic protocol separate from the turn-level evaluator assigns every executed non-passing call to one of the
six mechanisms using current- and previous-turn context, screenshots, source/DOM evidence, actor/tool traces, and
build or output evidence where applicable. These labels are diagnostic and do not affect official pass/fail
metrics. Three non-author annotators, blinded to judge labels and model identities, independently label a
category-balanced sample of 120 failures (20 per category) to evaluate agreement with the diagnostic judge.

\FloatBarrier
% Restore the standard double-column float-page glue for later appendices.
\makeatletter
\setlength{\@dblfptop}{0pt plus 1fil}
\setlength{\@dblfpsep}{8pt plus 2fil}
\setlength{\@dblfpbot}{0pt plus 1fil}
\makeatother
\setcounter{dbltopnumber}{2}
\setcounter{totalnumber}{3}
\renewcommand{\dbltopfraction}{0.7}
\renewcommand{\textfraction}{0.2}

\section{Qualitative Analysis and Visualization Tools}

\subsection{Web-based Visualization Interface}
\label{appendix:web-visualization}

\benchmark{} includes a local interface for inspecting benchmark composition and run artifacts.

\begin{figure}[tbp]
    \centering
    \includegraphics[width=0.84\linewidth,trim=1bp 0 0 0,clip]{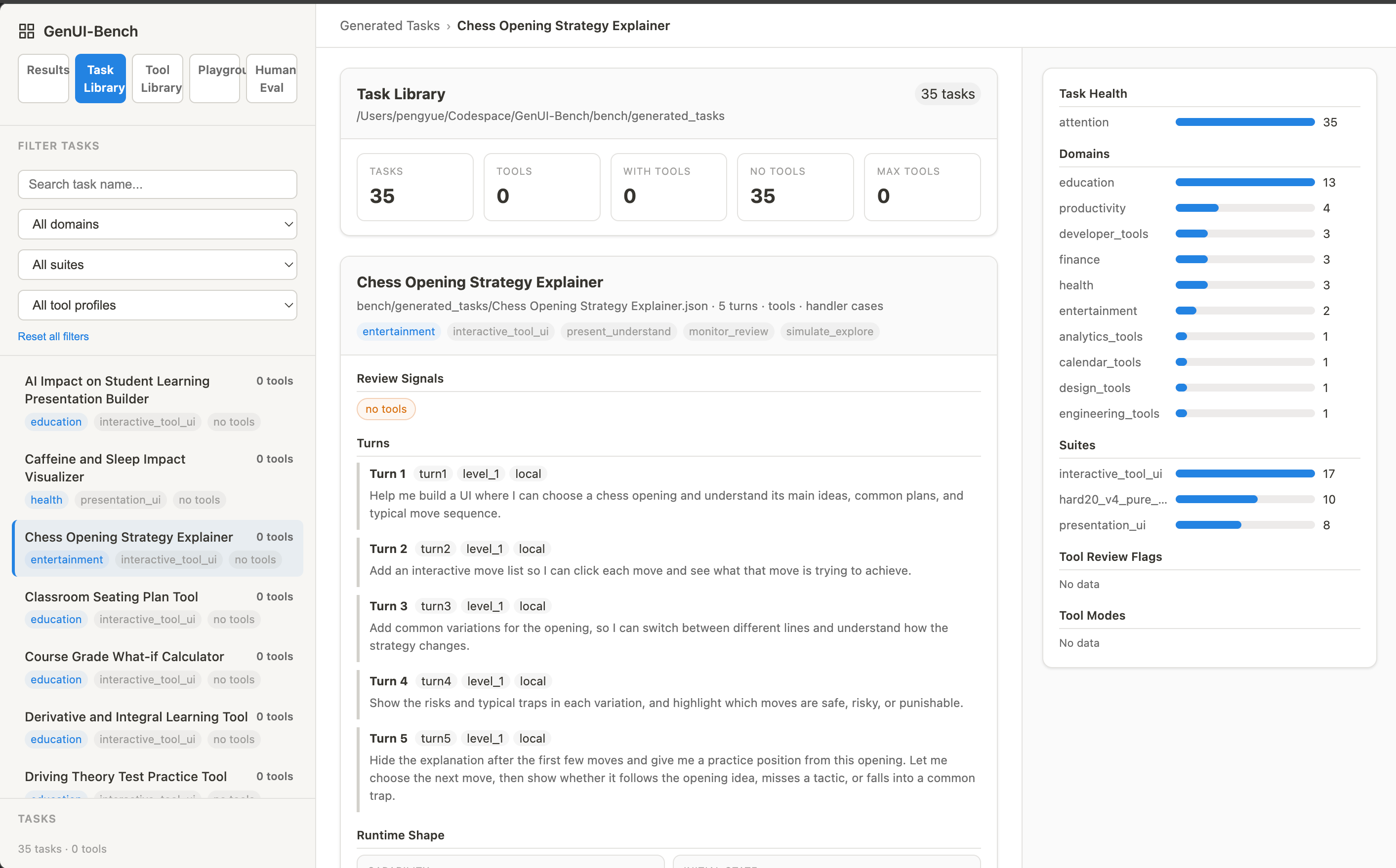}
    \captionsetup{hypcap=false}
    \caption{\textbf{Task library.} The web interface summarizes task metadata, domain distribution, suite membership, tool availability, and user turns.}
    \label{fig:web-task-library}
\end{figure}

The results view connects aggregate scores to the evidence for a selected turn. A manual audit can inspect the
request, source, screenshot, actor trace, tool logs, backend state, and evaluator output in one record.

\begin{figure}[tbp]
    \centering
    \includegraphics[width=0.94\linewidth,trim=0 0 0 8bp,clip]{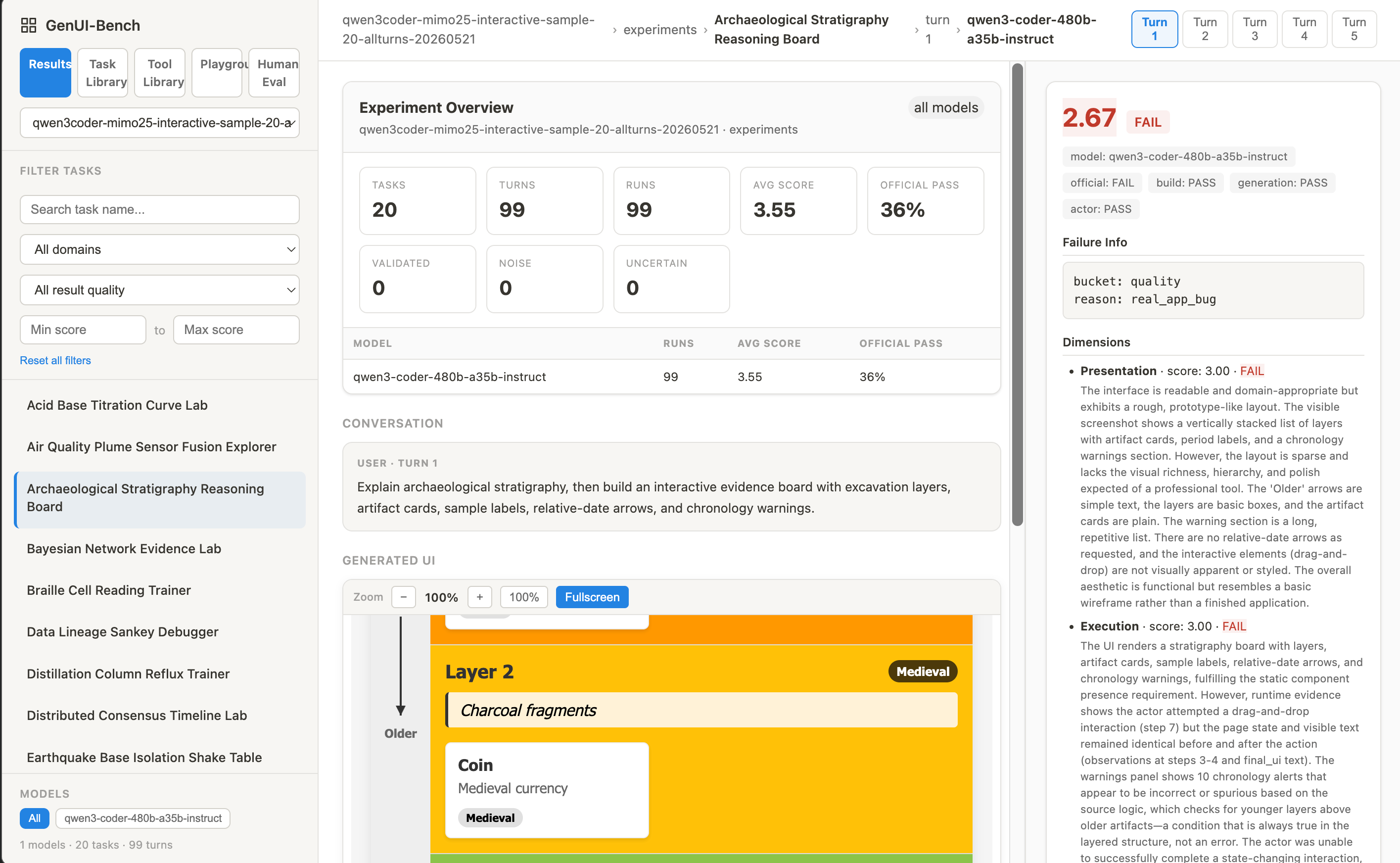}
    \caption{\textbf{Experiment results.} Aggregate scores link to the generated UI, execution status, actor result, and dimension-level evaluator feedback.}
    \label{fig:web-results-view}
\end{figure}

\FloatBarrier

\section{Prompt Templates}
\label{app:prompts}

\subsection{Generator Prompt}
\label{app:generator-prompt}

Every model receives the same system prompt.

\begin{promptbox}{Generator prompt excerpt}
You are an LLM chatbot with the ability to build a small, self-contained React app in a sandboxed workspace when it is helpful. Output \texttt{assistant\_text:} followed by a non-empty user-facing response, then complete source file sections. Treat the app as one accumulated artifact across turns. Implement the current request while preserving prior behavior that is still relevant. For tool-backed workflows, use only declared \texttt{callTool} and \texttt{readResource} capabilities, show returned results and committed readback, and do not fake tool results.
\end{promptbox}

\subsection{Turn Payload Template}
\label{app:turn-payload}

At turn $t$, the payload combines the current request and public task contracts with compact prior-turn context and
the latest source files.

\begin{jsonlistingbox}{Turn Payload Template}
{
  "current_user_request": "...",
  "task": {
    "context": "...",
    "tools": [... public contracts ...],
    "resources": [... public contracts ...],
    "conversation_history": [...]
  },
  "previous_turns": [
    {
      "turn": 1,
      "user_request": "...",
      "assistant_text": "...",
      "final_ui_text": "..."
    }
  ],
  "previous_turn_source": {
    "turn": 1,
    "files": {
      "src/App.tsx": "...",
      "src/App.css": "..."
    }
  }
}
\end{jsonlistingbox}

% Balance the final appendix page across both columns.
\newpage
\subsection{Browser Actor Prompt}
\label{app:actor-prompt}

\begin{promptbox}{Browser actor prompt excerpt}
Open the generated app and evaluate whether it satisfies the current user request. Prioritize the current turn's request over exhaustive testing. Identify the primary action, execute it, inspect the result, and finish once decisive evidence is available. Runtime logs show which tool calls actually happened; use them to verify behavior.
\end{promptbox}

\subsection{Evaluator Prompt}
\label{app:evaluator-prompt}

\begin{promptbox}{Evaluator prompt excerpt}
You are the \benchmark{} evaluator. Grade generated UI code based strictly on observable evidence. Use only the supplied evidence. Source code is not runtime proof. Return exactly one JSON object with one top-level key per requested dimension; each value must contain \texttt{score}, \texttt{summary}, and \texttt{failure\_types}.
\end{promptbox}

\end{document}